\documentclass{article}

     \usepackage[preprint]{neurips_2022}

\usepackage[utf8]{inputenc} 
\usepackage{bm}             
\usepackage[T1]{fontenc}    
\usepackage{hyperref}       
\usepackage{url}            
\usepackage{booktabs}       
\usepackage{amsfonts}       
\usepackage{amsmath}        
\usepackage{nicefrac}       
\usepackage{microtype}      
\usepackage{xcolor}         
\usepackage[pdftex]{graphicx}
\usepackage[inline]{enumitem} 
\usepackage{caption}        
\usepackage{subcaption}     

\usepackage{fancyhdr}
\usepackage{mathtools}
\usepackage{dsfont}
\usepackage{algorithm}
\usepackage{algpseudocode}

\usepackage{version}
\excludeversion{kurz}
\includeversion{tune}
\excludeversion{fig_publish}
\includeversion{fig_count_words}

\setcitestyle{authoryear}
\title{Automated wildlife image classification: An active learning tool for ecological applications}

\author{%
  Ludwig Bothmann$^{a,b,*}$, Lisa Wimmer$^{a,b}$, Omid Charrakh$^c$, Tobias Weber$^{a,b}$, 
  \And Hendrik Edelhoff$^d$, Wibke Peters$^{d,e}$, Hien Nguyen$^d$, 
  \And Caryl Benjamin$^{f}$, Annette Menzel$^{f,g}$
}

\begin{document}

\ifdefined\N                                                                
\renewcommand{\N}{\mathds{N}} 
\else \newcommand{\N}{\mathds{N}} \fi 
\newcommand{\Z}{\mathds{Z}} 
\newcommand{\Q}{\mathds{Q}} 
\newcommand{\R}{\mathds{R}} 
\ifdefined\C 
  \renewcommand{\C}{\mathds{C}} 
\else \newcommand{\C}{\mathds{C}} \fi
\newcommand{\continuous}{\mathcal{C}} 
\newcommand{\M}{\mathcal{M}} 
\newcommand{\epsm}{\epsilon_m} 

\newcommand{\setzo}{\{0, 1\}} 
\newcommand{\setmp}{\{-1, +1\}} 
\newcommand{\unitint}{[0, 1]} 

\newcommand{\xt}{\tilde x} 
\newcommand{\argmax}{\operatorname{arg\,max}} 
\newcommand{\argmin}{\operatorname{arg\,min}} 
\newcommand{\argminlim}{\mathop{\mathrm{arg\,min}}\limits} 
\newcommand{\argmaxlim}{\mathop{\mathrm{arg\,max}}\limits} 
\newcommand{\sign}{\operatorname{sign}} 
\newcommand{\I}{\mathbb{I}} 
\newcommand{\order}{\mathcal{O}} 
\newcommand{\pd}[2]{\frac{\partial{#1}}{\partial #2}} 
\newcommand{\floorlr}[1]{\left\lfloor #1 \right\rfloor} 
\newcommand{\ceillr}[1]{\left\lceil #1 \right\rceil} 

\newcommand{\sumin}{\sum\limits_{i=1}^n} 
\newcommand{\sumim}{\sum\limits_{i=1}^m} 
\newcommand{\sumjn}{\sum\limits_{j=1}^n} 
\newcommand{\sumjp}{\sum\limits_{j=1}^p} 
\newcommand{\sumik}{\sum\limits_{i=1}^k} 
\newcommand{\sumkg}{\sum\limits_{k=1}^g} 
\newcommand{\sumjg}{\sum\limits_{j=1}^g} 
\newcommand{\meanin}{\frac{1}{n} \sum\limits_{i=1}^n} 
\newcommand{\meanim}{\frac{1}{m} \sum\limits_{i=1}^m} 
\newcommand{\meankg}{\frac{1}{g} \sum\limits_{k=1}^g} 
\newcommand{\prodin}{\prod\limits_{i=1}^n} 
\newcommand{\prodkg}{\prod\limits_{k=1}^g} 
\newcommand{\prodjp}{\prod\limits_{j=1}^p} 

\newcommand{\one}{\boldsymbol{1}} 
\newcommand{\zero}{\mathbf{0}} 
\newcommand{\id}{\boldsymbol{I}} 
\newcommand{\diag}{\operatorname{diag}} 
\newcommand{\trace}{\operatorname{tr}} 
\newcommand{\spn}{\operatorname{span}} 
\newcommand{\scp}[2]{\left\langle #1, #2 \right\rangle} 
\newcommand{\mat}[1]{\begin{pmatrix} #1 \end{pmatrix}} 
\newcommand{\Amat}{\mathbf{A}} 
\newcommand{\Deltab}{\mathbf{\Delta}} 

\renewcommand{\P}{\mathds{P}} 
\newcommand{\E}{\mathds{E}} 
\newcommand{\var}{\mathsf{Var}} 
\newcommand{\cov}{\mathsf{Cov}} 
\newcommand{\corr}{\mathsf{Corr}} 
\newcommand{\normal}{\mathcal{N}} 
\newcommand{\iid}{\overset{i.i.d}{\sim}} 
\newcommand{\distas}[1]{\overset{#1}{\sim}} 

\newcommand{\Xspace}{\mathcal{X}} 
\newcommand{\Yspace}{\mathcal{Y}} 
\newcommand{\nset}{\{1, \ldots, n\}} 
\newcommand{\pset}{\{1, \ldots, p\}} 
\newcommand{\gset}{\{1, \ldots, g\}} 
\newcommand{\Pxy}{\mathbb{P}_{xy}} 
\newcommand{\Exy}{\mathbb{E}_{xy}} 
\newcommand{\xv}{\mathbf{x}} 
\newcommand{\xtil}{\tilde{\mathbf{x}}} 
\newcommand{\yv}{\mathbf{y}} 
\newcommand{\xy}{(\xv, y)} 
\newcommand{\xvec}{\left(x_1, \ldots, x_p\right)^T} 
\newcommand{\Xmat}{\mathbf{X}} 
\newcommand{\allDatasets}{\mathds{D}} 
\newcommand{\allDatasetsn}{\mathds{D}_n}  
\newcommand{\D}{\mathcal{D}} 
\newcommand{\Dn}{\D_n} 
\newcommand{\Dtrain}{\mathcal{D}_{\text{train}}} 
\newcommand{\Dtest}{\mathcal{D}_{\text{test}}} 
\newcommand{\xyi}[1][i]{\left(\xv^{(#1)}, y^{(#1)}\right)} 
\newcommand{\Dset}{\left( \xyi[1], \ldots, \xyi[n]\right)} 
\newcommand{\defAllDatasetsn}{(\Xspace \times \Yspace)^n} 
\newcommand{\defAllDatasets}{\bigcup_{n \in \N}(\Xspace \times \Yspace)^n} 
\newcommand{\xdat}{\left\{ \xv^{(1)}, \ldots, \xv^{(n)}\right\}} 
\newcommand{\yvec}{\left(y^{(1)}, \hdots, y^{(n)}\right)^T} 
\renewcommand{\xi}[1][i]{\xv^{(#1)}} 
\newcommand{\yi}[1][i]{y^{(#1)}} 
\newcommand{\xivec}{\left(x^{(i)}_1, \ldots, x^{(i)}_p\right)^T} 
\newcommand{\xj}{\xv_j} 
\newcommand{\xjvec}{\left(x^{(1)}_j, \ldots, x^{(n)}_j\right)^T} 
\newcommand{\phiv}{\mathbf{\phi}} 
\newcommand{\phixi}{\mathbf{\phi}^{(i)}} 

\newcommand{\lamv}{\bm{\lambda}} 
\newcommand{\Lam}{\bm{\Lambda}}	 
\newcommand{\preimageInducer}{\left(\defAllDatasets\right)\times\Lam} 
\newcommand{\preimageInducerShort}{\allDatasets\times\Lam} 
\newcommand{\ind}{\mathcal{I}} 

\newcommand{\ftrue}{f_{\text{true}}}  
\newcommand{\ftruex}{\ftrue(\xv)} 
\newcommand{\fx}{f(\xv)} 
\newcommand{\fdomains}{f: \Xspace \rightarrow \R^g} 
\newcommand{\Hspace}{\mathcal{H}} 
\newcommand{\fbayes}{f^{\ast}} 
\newcommand{\fxbayes}{f^{\ast}(\xv)} 
\newcommand{\fkx}[1][k]{f_{#1}(\xv)} 
\newcommand{\fh}{\hat{f}} 
\newcommand{\fxh}{\fh(\xv)} 
\newcommand{\fxt}{f(\xv ~|~ \thetab)} 
\newcommand{\fxi}{f\left(\xv^{(i)}\right)} 
\newcommand{\fxih}{\hat{f}\left(\xv^{(i)}\right)} 
\newcommand{\fxit}{f\left(\xv^{(i)} ~|~ \thetab\right)} 
\newcommand{\fhD}{\fh_{\D}} 
\newcommand{\fhDtrain}{\fh_{\Dtrain}} 
\newcommand{\fhDnlam}{\fh_{\Dn, \lamv}} 
\newcommand{\fhDlam}{\fh_{\D, \lamv}} 
\newcommand{\fhDnlams}{\fh_{\Dn, \lamv^\ast}} 
\newcommand{\fhDlams}{\fh_{\D, \lamv^\ast}} 

\newcommand{\hx}{h(\xv)} 
\newcommand{\hh}{\hat{h}} 
\newcommand{\hxh}{\hat{h}(\xv)} 
\newcommand{\hxt}{h(\xv | \thetab)} 
\newcommand{\hxi}{h\left(\xi\right)} 
\newcommand{\hxit}{h\left(\xi ~|~ \thetab\right)} 
\newcommand{\hbayes}{h^{\ast}} 
\newcommand{\hxbayes}{h^{\ast}(\xv)} 

\newcommand{\yh}{\hat{y}} 
\newcommand{\yih}{\hat{y}^{(i)}} 
\newcommand{\resi}{\yi- \yih}

\newcommand{\thetah}{\hat{\theta}} 
\newcommand{\thetab}{\bm{\theta}} 
\newcommand{\thetabh}{\bm{\hat\theta}} 
\newcommand{\thetat}[1][t]{\thetab^{[#1]}} 
\newcommand{\thetatn}[1][t]{\thetab^{[#1 +1]}} 
\newcommand{\thetahDnlam}{\thetabh_{\Dn, \lamv}} 
\newcommand{\thetahDlam}{\thetabh_{\D, \lamv}} 
\newcommand{\mint}{\min_{\thetab \in \Theta}} 
\newcommand{\argmint}{\argmin_{\thetab \in \Theta}} 

\newcommand{\pdf}{p} 
\newcommand{\pdfx}{p(\xv)} 
\newcommand{\pixt}{\pi(\xv~|~ \thetab)} 
\newcommand{\pixit}{\pi\left(\xi ~|~ \thetab\right)} 
\newcommand{\pixii}{\pi(\xi)} 
\newcommand{\pii}{\pi^{(i)}} 

\newcommand{\pdfxy}{p(\xv,y)} 
\newcommand{\pdfxyt}{p(\xv, y ~|~ \thetab)} 
\newcommand{\pdfxyit}{p\left(\xi, \yi ~|~ \thetab\right)} 

\newcommand{\pdfxyk}[1][k]{p(\xv | y= #1)} 
\newcommand{\lpdfxyk}[1][k]{\log p(\xv | y= #1)} 
\newcommand{\pdfxiyk}[1][k]{p\left(\xi | y= #1 \right)} 

\newcommand{\pik}[1][k]{\pi_{#1}} 
\newcommand{\lpik}[1][k]{\log \pi_{#1}} 
\newcommand{\pit}{\pi(\thetab)} 

\newcommand{\post}{\P(y = 1 ~|~ \xv)} 
\newcommand{\postk}[1][k]{\P(y = #1 ~|~ \xv)} 
\newcommand{\pidomains}{\pi: \Xspace \rightarrow \unitint} 
\newcommand{\pibayes}{\pi^{\ast}} 
\newcommand{\pixbayes}{\pi^{\ast}(\xv)} 
\newcommand{\pix}{\pi(\xv)} 
\newcommand{\pikx}[1][k]{\pi_{#1}(\xv)} 
\newcommand{\pikxt}[1][k]{\pi_{#1}(\xv ~|~ \thetab)} 
\newcommand{\pixh}{\hat \pi(\xv)} 
\newcommand{\pikxh}[1][k]{\hat \pi_{#1}(\xv)} 
\newcommand{\pixih}{\hat \pi(\xi)} 
\newcommand{\pikxih}[1][k]{\hat \pi_{#1}(\xi)} 
\newcommand{\pdfygxt}{p(y ~|~\xv, \thetab)} 
\newcommand{\pdfyigxit}{p\left(\yi ~|~\xi, \thetab\right)} 
\newcommand{\lpdfygxt}{\log \pdfygxt } 
\newcommand{\lpdfyigxit}{\log \pdfyigxit} 

\newcommand{\bayesrulek}[1][k]{\frac{\P(\xv | y= #1) \P(y= #1)}{\P(\xv)}} 
\newcommand{\muk}{\bm{\mu_k}} 

\newcommand{\eps}{\epsilon} 
\newcommand{\epsi}{\epsilon^{(i)}} 
\newcommand{\epsh}{\hat{\epsilon}} 
\newcommand{\yf}{y \fx} 
\newcommand{\yfi}{\yi \fxi} 
\newcommand{\Sigmah}{\hat \Sigma} 
\newcommand{\Sigmahj}{\hat \Sigma_j} 

\newcommand{\Lyf}{L\left(y, f\right)} 
\newcommand{\Lxy}{L\left(y, \fx\right)} 
\newcommand{\Lxyi}{L\left(\yi, \fxi\right)} 
\newcommand{\Lxyt}{L\left(y, \fxt\right)} 
\newcommand{\Lxyit}{L\left(\yi, \fxit\right)} 
\newcommand{\Lxym}{L\left(\yi, f\left(\bm{\tilde{x}}^{(i)} ~|~ \thetab\right)\right)} 
\newcommand{\Lpixy}{L\left(y, \pix\right)} 
\newcommand{\Lpixyi}{L\left(\yi, \pixii\right)} 
\newcommand{\Lpixyt}{L\left(y, \pixt\right)} 
\newcommand{\Lpixyit}{L\left(\yi, \pixit\right)} 
\newcommand{\Lhxy}{L\left(y, \hx\right)} 
\newcommand{\Lr}{L\left(r\right)} 
\newcommand{\lone}{|y - \fx|} 
\newcommand{\ltwo}{\left(y - \fx\right)^2} 
\newcommand{\lbernoullimp}{\ln(1 + \exp(-y \cdot \fx))} 
\newcommand{\lbernoullizo}{- y \cdot \fx + \log(1 + \exp(\fx))} 
\newcommand{\lcrossent}{- y \log \left(\pix\right) - (1 - y) \log \left(1 - \pix\right)} 
\newcommand{\lbrier}{\left(\pix - y \right)^2} 
\newcommand{\risk}{\mathcal{R}} 
\newcommand{\riskbayes}{\mathcal{R}^\ast}
\newcommand{\riskf}{\risk(f)} 
\newcommand{\riskdef}{\E_{y|\xv}\left(\Lxy \right)} 
\newcommand{\riskt}{\mathcal{R}(\thetab)} 
\newcommand{\riske}{\mathcal{R}_{\text{emp}}} 
\newcommand{\riskeb}{\bar{\mathcal{R}}_{\text{emp}}} 
\newcommand{\riskef}{\riske(f)} 
\newcommand{\risket}{\mathcal{R}_{\text{emp}}(\thetab)} 
\newcommand{\riskr}{\mathcal{R}_{\text{reg}}} 
\newcommand{\riskrt}{\mathcal{R}_{\text{reg}}(\thetab)} 
\newcommand{\riskrf}{\riskr(f)} 
\newcommand{\riskrth}{\hat{\mathcal{R}}_{\text{reg}}(\thetab)} 
\newcommand{\risketh}{\hat{\mathcal{R}}_{\text{emp}}(\thetab)} 
\newcommand{\LL}{\mathcal{L}} 
\newcommand{\LLt}{\mathcal{L}(\thetab)} 
\newcommand{\LLtx}{\mathcal{L}(\thetab | \xv)} 
\newcommand{\logl}{\ell} 
\newcommand{\loglt}{\logl(\thetab)} 
\newcommand{\logltx}{\logl(\thetab | \xv)} 
\newcommand{\errtrain}{\text{err}_{\text{train}}} 
\newcommand{\errtest}{\text{err}_{\text{test}}} 
\newcommand{\errexp}{\overline{\text{err}_{\text{test}}}} 

\newcommand{\thx}{\thetab^T \xv} 
\newcommand{\olsest}{(\Xmat^T \Xmat)^{-1} \Xmat^T \yv} 

\newcommand{\lb}[1]{\textcolor{blue}{[#1]}}
\newcommand{\bp}[1]{\textcolor{blue}{$\rightarrow$ #1 \\}}

\newcommand{\myemph}[1]{\textbf{#1}}

\newcommand{\titel}{
Combining object detection, transfer learning and active learning improves out-of-sample prediction in small wildlife image data sets}

\newcommand{\pih}{\hat{\pi}}
\newcommand{\pihf}{\hat{\pi}(\cdot)}
\newcommand{\Dtune}{\D_{tune}}
\newcommand{\Dval}{\D_{val}}
\newcommand{\DTrain}{\D_{train}}
\newcommand{\DTest}{\D_{test}}

\include{ml-svm}

\maketitle
$^a$Department of Statistics, LMU Munich, Ludwigstr. 33, 80539 München, Germany\\
$^b$Munich Center for Machine Learning (MCML), Ludwigstr. 33, 80539 München, Germany\\
$^c$Munich Center for Mathematical Philosophy (MCMP), Ludwigstr. 31, 80539 München, LMU Munich, Germany\\
$^d$Wildlife Biology and Management Research Unit, Bavarian State Institute of Forestry (LWF), Hans-Carl-von-Carlowitz-Platz 1, 85354 Freising, Germany\\
$^e$Wildlife Biology and Management Unit, Technical University of Munich, Hans-Carl-von-Carlowitz-Platz 2, 85354 Freising, Germany\\
$^f$Ecoclimatology, TUM School of Life Sciences, Technical University of Munich, Hans-Carl-von-Carlowitz-Platz 2, 85354 Freising, Germany\\
$^g$TUM Institute for Advanced Study, Lichtenbergstraße 2 a
85748 Garching, Germany\\
$^*$Corresponding author: \texttt{ludwig.bothmann@stat.uni-muenchen.de}\\[2ex]

\begin{abstract}

Wildlife camera trap images are being used extensively to investigate animal abundance, habitat associations, and behavior, which is complicated by the fact that experts must first classify the images to retrieve relevant information. 
Artificial intelligence systems can take over this task but usually need a large number of already-labeled training images to achieve sufficient performance.
This requirement necessitates human expert labor and poses a particular challenge for projects with few cameras or short durations.
We propose a label-efficient learning strategy that enables researchers with small or medium-sized image databases to leverage the potential of modern machine learning, thus freeing crucial resources for subsequent analyses.\\
Our methodological proposal is two-fold: On the one hand, we improve current strategies of combining object detection and image classification by tuning the hyperparameters of both models. On the other hand, we provide an active learning system that allows training deep learning models very efficiently in terms of required manually labeled training images. We supply a software package that enables researchers to use these methods without specific programming skills and thereby ensure the broad applicability of the proposed framework in ecological practice.\\
We show that our tuning strategy improves predictive performance, emphasizing that tuning can and must be done separately for a new data set. We demonstrate how the active learning pipeline reduces the amount of pre-labeled data needed to achieve specific predictive performance and that it is especially valuable for improving out-of-sample predictive performance.\\
We conclude that the combination of tuning and active learning increases the predictive performance of automated image classifiers substantially. Furthermore, we argue that our work can broadly impact the community through the ready-to-use software package provided. Finally, the publication of our models tailored to European wildlife data enriches existing model bases mostly trained on data from Africa and North America.

\end{abstract}


\section{Introduction}
\label{sec:intro}

Wildlife camera traps have been increasingly used for estimating the abundance and distribution of animal communities, studying animal behavior, and assessing animal-plant interactions over the last decade 
\citep{delisle_next-generation_2021, trolliet_use_2014, tuia_perspectives_2022}. 
Since manual annotation of large image data sets is time-consuming and expensive, machine learning (ML) techniques, especially from the field of deep learning (DL) 
have been developed, customized, and applied to wildlife image data sets 
\citep{auer_minimizing_2021, beery_efficient_2019, christin_applications_2019, gimenez_trade-off_2021,  miao_iterative_2021, norouzzadeh_automatically_2018, norouzzadeh_deep_2021, schneider_past_2019, schneider_three_2020,tabak_machine_2019, tabak_improving_2020, tabak_cameratrapdetector_2022, whytock_robust_2021}. 
These techniques allow for high detection rates in the investigated data sets. 

However, the broad applicability of the proposed methods is hindered for different reasons: First, models are often trained on data sets from specific regions of the earth, e.g., the Serengeti Snapshot data from Africa \citep{swanson_data_2015}, the North American Camera Trap Images  \citep{tabak_machine_2019} and Caltech Camera Traps \citep{beery_recognition_2018} from North America, the LILA database\footnote{https://lila.science/datasets} with images from outside Europe, and DeepFaune \citep{rigoudy_deepfaune_2022} from France. Applying such models to other animal species or habitats, which are not representative of the data used for model training, leads to a drop in  
predictive performance, 
see also \cite{beery_recognition_2018, koh_wilds_2021, schneider_three_2020, shepley_automated_2021, tuia_perspectives_2022}. Second, training such models for new animal species or new habitats demands a large amount of manually annotated training images, which may not be available for smaller research groups that operate in regions where models trained on the above-mentioned data sets result in poor classification performance.
Third, highly customizable software targeted at experienced programmers may be an obstacle for applied researchers from ecology who would benefit much more from an easy-to-use software package that yields high-performing ML models across a range of tasks, see, e.g., \cite{velez_choosing_2022, gimenez_trade-off_2021}.

We aim at filling this gap by proposing an active learning pipeline that yields high-performing models, demands only small training data sets, and is readily usable -- via accompanying example code and tutorials -- for researchers in ecology without any specific background in ML or computer science. Our main research goals are:

\begin{itemize}
    \item achieving high out-of-sample (OOS) performance, i.e., good transferability to new domains,
    \item dealing with both existing and new, previously unseen animal species,
    \item providing solutions for small data sets ($< 50.000$ images),
    \item adding models pre-trained on European wildlife images to the existing pre-trained models that are mainly from Africa and North America, and 
    \item supplying highly usable software that can be used off-the-shelf at a low computational cost.
\end{itemize}

\subsection{Main contributions}

\subsubsection{Active learning pipeline}

Our pipeline consists of four main building blocks: We combine the benefits of \myemph{object detection} -- to locate parts of an image that most likely contain animals -- and \myemph{image classification} -- to classify the animal species found by the object detector. A thorough \myemph{tuning process} optimizes the hyperparameters of both the object detector and image classifier, including the architecture of the deep neural network used for transfer learning. Finally, an \myemph{active learning} component allows training a new model or adapting a pre-trained model to new animal classes or habitats with a substantially reduced need for training data from that new domain. We will explain the central components of the pipeline in the following and refer to ``Materials and Methods'' for a thorough explanation of all steps.

\paragraph{Combining object detection and image classification}

A key factor for creating a powerful pipeline is the combination of object detection and image classification, see also \cite{curry_application_2021}. 
First, using a pre-trained object detection network, as proposed, e.g., by \cite{beery_efficient_2019}, allows for rapid identification of empty images \citep{velez_choosing_2022}.  
Otherwise, the task of locating empty images is either very time-consuming --  \cite{gimenez_trade-off_2021} mention that doing this manually ``took several weeks of labor full time'' for a rather small data set of $<50.000$ images --
or technically complicated without yielding convincing results (e.g., \cite{auer_minimizing_2021} tried to use active learning to filter out empty images).
Second, image backgrounds and camera-added information (such as a header or footer) are removed and the image classifier can directly focus on the parts of the image where the object detector has detected an animal.
As a result, fewer training images are necessary to adapt pre-trained models to new domains; furthermore, the number of animals on a given image can be estimated, which, in turn, is desirable from an ecological perspective for wildlife abundance estimation methods \citep{rowcliffe_estimating_2008, royle_n-mixture_2004, moeller_three_2018}.

We enrich the pipeline with the following features:

\begin{itemize}
    \item \textbf{Tuned confidence threshold for object detection:} When using an object detection method such as the \textit{Megadetector} (MD) \citep{beery_efficient_2019}, each predicted bounding box for class ``animal'' is accompanied by a confidence score.
While others use a fixed confidence threshold above which an image is deemed non-empty (e.g., $0.9$ as in \cite{norouzzadeh_deep_2021}), we consider this optimal threshold a performance-relevant hyperparameter and propose to tune it. Thereby, we optimize the trade-off between (i) ending up with many empty images left after object detection (i.e.,  threshold for class ``animal'' is too low) and (ii) overlooking animal images (i.e., threshold for class ``animal'' is too high).

\item \textbf{Augmentation:} Data augmentation, i.e., enlarging the training set by random modifications of the existing samples, is known to improve results of image classifiers \citep{shorten_survey_2019}. 
We include this step such as \cite{norouzzadeh_automatically_2018, schneider_three_2020, tabak_machine_2019, whytock_robust_2021} -- in contrast to other proposed pipelines such as \cite{norouzzadeh_deep_2021, tabak_cameratrapdetector_2022, tabak_improving_2020}.

\item \textbf{Transfer learning:} Transfer learning \citep{tan_survey_2018} 
refers to a technique where large pre-trained models are fine-tuned on the data at hand, hence leveraging universal knowledge these models have gained from learning on millions of images before. 
We support transfer learning with different models  
pre-trained on the ImageNet database \citep{deng_imagenet_2009} that consists of around 1.3 million images from 1.000 classes and consider Xception \citep{chollet_xception_2017}, InceptionResNetV2 \citep{szegedy_inception-v4_2016}, and DenseNet121 
\citep{huang_densely_2017} 
in this work; other such pre-trained networks can be easily included using our software.

\item \textbf{Hyperparameter tuning:} Hyperparameter tuning is a common technique to optimize the hyperparameters of an ML model \citep{bischl_hyperparameter_2023}. 
However, the adoption of hyperparameter tuning in the field of DL is still in its infancy \citep{ottoni_hyperparameter_2022, hutter_automated_2019} due to the higher computational burden compared to smaller tabular data sets. 
We argue that this cost is worth bearing in light of the expected performance gains. 
Besides the threshold of the bounding-box confidence, we optimize the choice of the best pre-trained model for the data at hand, which further improves the results.

\end{itemize}

\paragraph{Active learning}

Adapting an existing model to a new task -- be it a new habitat, new animal species, or new classes (as in moving from ImageNet objects to animals during transfer learning) -- involves manually labeling data and re-training the model with this labeled data. A practical problem is that we do not know beforehand how many and which images should be labeled manually in order to achieve a certain predictive performance. Active learning (AL) \citep{settles_active_2009, yang_active_2018} tackles both challenges at the same time: The model is trained with a human-in-the-loop who sequentially labels batches of data and terminates the process when the model has reached sufficient predictive performance.
In each iteration, the AL system requests labels for the images it deems most informative for its learning process.
This careful iterative data selection reduces the amount of required human labels considerably, as our results in Section~\ref{sec:oosample} confirm.

\subsubsection{Software package}

The presented pipeline can be used from different perspectives, depending on the use case at hand and the desired level of customizability. All of them allow training models for new domains and new animal species.

\begin{itemize}
    \item \textbf{AL for everybody:}
For users who want to train a model on their wildlife image data in order to automatically label large parts of their image database, we offer a command line interface (CLI). With this CLI, users can iterate the AL loop until a satisfying predictive performance is reached, and eventually predict the class labels of all remaining unlabeled images. No programming skills are required. We provide a step-by-step example of how to use the CLI.\footnote{https://github.com/slds-lmu/wildlife-experiments, also includes code for reproducing our results}

\item \textbf{Full customizability:}
Users who like to have full flexibility and be able to customize the code base for their more specific needs may access the Python code base directly, which uses Keras \citep{chollet2015keras} and TensorFlow \citep{abadi_tensorflow_2015} as backend.\footnote{https://github.com/slds-lmu/wildlife-ml}

\end{itemize}

\subsubsection{Pre-trained model for European wildlife images}

We trained the pipeline on a set of European wildlife images classified into the six most abundant species (European hare (\textit{Lepus europaeus}), red deer (\textit{Cervus elaphus}), red fox (\textit{Vulpes vulpes}), red squirrel (\textit{Sciurus vulgaris}), roe deer (\textit{Capreolus capreolus}) and wild boar (\textit{Sus scrofa})) detected in camera traps in the state of Bavaria in southeastern Germany. The training data additionally includes a class ``empty'' as well as a class ``others'' to account for other species not categorized yet.  
The trained models are publicly available\footnote{Weights for the neural network can be found here: \url{https://syncandshare.lrz.de/getlink/fiJsgDEKtkLCXfbhWM1GLR/ckpt_final_model.hdf5}} and interested researchers can use those either as-is or as a starting point for fine-tuning them with the AL pipeline implemented in our software package.

\subsection{Related work}

For over a decade, computer vision methods are used to analyze and classify wildlife camera trap images. While early contributions use pattern matching \citep{bolger_computer-assisted_2012} or feature extraction followed by a classification via support vector machines \citep{yu_automated_2013}, \cite{chen_deep_2014} introduced using convolutional neural networks (CNNs) and an early form of object detection to the wildlife camera trap literature. \cite{gomez_villa_towards_2017} introduce transfer learning to improve the performance of CNN classification.
Using DL methods for the automatic classification of wildlife camera trap images has widely spread in the following years, the work of \cite{norouzzadeh_automatically_2018} being another key contribution. Subsequently, several research groups incorporated an object detection component, e.g., \cite{norouzzadeh_deep_2021, shepley_automated_2021, tabak_cameratrapdetector_2022}
or discussed OOS performance, e.g., \cite{auer_minimizing_2021, curry_application_2021, gimenez_trade-off_2021, miao_iterative_2021, schneider_three_2020, shepley_automated_2021, whytock_robust_2021, tabak_improving_2020}. 
Active learning was used for classifying wildlife images from drones by \cite{kellenberger_half_2019} and proposed for camera trap images by, e.g.,  \cite{auer_minimizing_2021, miao_iterative_2021, norouzzadeh_deep_2021}.
A variety of software frameworks have been proposed in the context of automated analysis of camera trap images; a recent discussion can be found in \cite{velez_choosing_2022}.

Perhaps the closest contribution to ours is \cite{norouzzadeh_deep_2021}, which also makes use of object detection, transfer learning, and active learning. However, we extend their pipeline and analysis by the following points: (1) We show empirically that an object detection component improves predictive performance -- compared to the simpler alternative of only using image classification; (2) we consider hyperparameter tuning and show the related improvements in predictive performance; (3) we support multiple transfer learning alternatives and decide on the best architecture in a data-driven manner during the tuning process; (4) we systematically analyze in-sample and out-of-sample errors for small data sets and quantify the benefit of using active learning; (5) we operate in a different and so far underrepresented domain -- European wildlife images -- and publish the pre-trained model.
Finally, (6) we provide ecologists with a software package and example code for Python, which enables users to actually apply the proposed methods without computational barriers.

Some work has been conducted on European wildlife images, e.g., \cite{rigoudy_deepfaune_2022, gimenez_trade-off_2021, auer_minimizing_2021} 
use image data from France and Germany, respectively. To the best of our knowledge, there has not been a contribution so far that resulted in a pre-trained model for European wildlife images and an active learning pipeline comparable to ours.

\section{Materials and methods}

\subsection{Data sets}
\label{sec:data}

\begin{figure}[t]
    \centering
    \includegraphics[width=\textwidth, trim=0 120 0 0, clip]{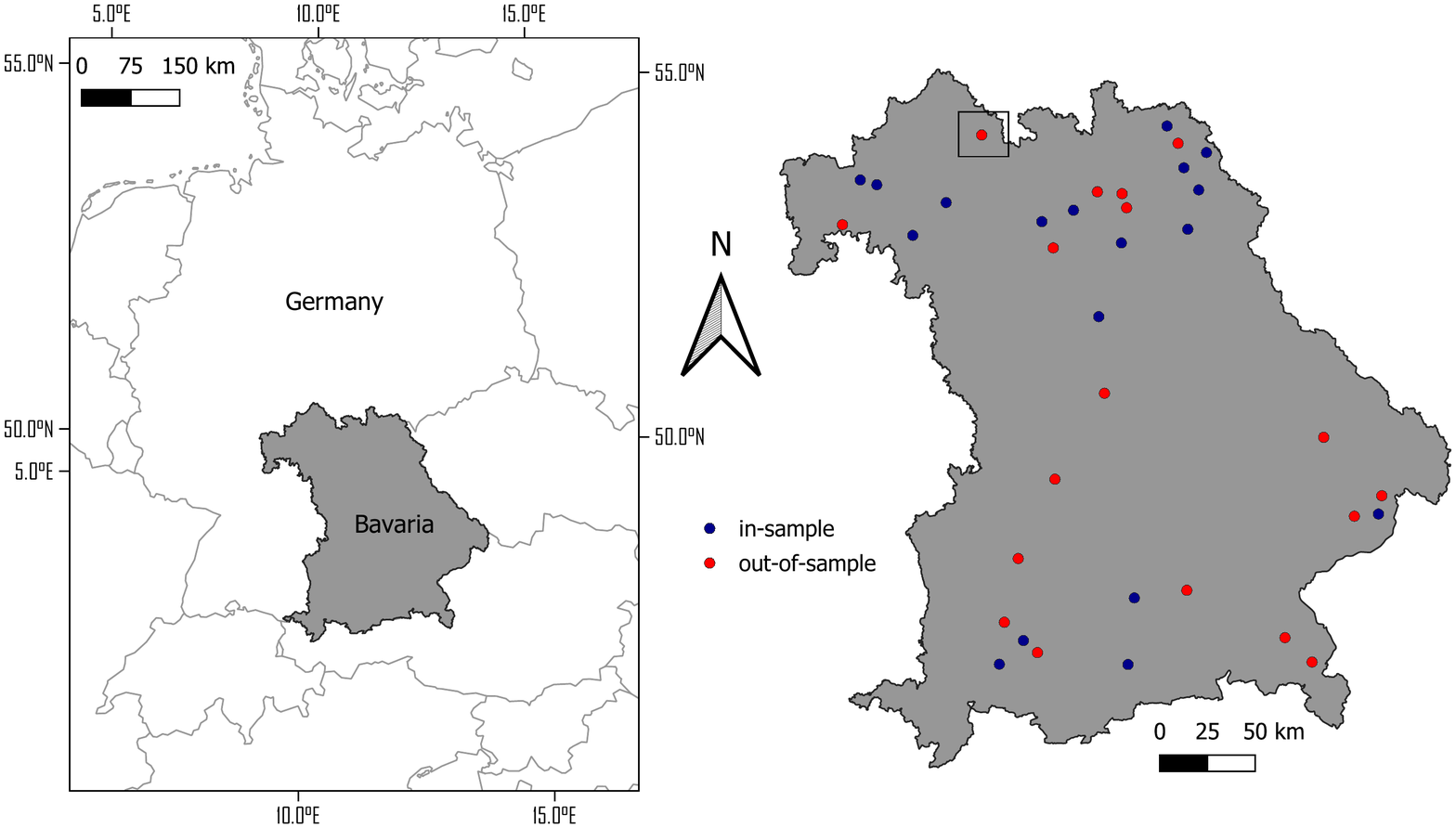}
    \caption{Study sites of the camera trap images used in this work. Two stations (black square) have almost the same location, the respective camera had to be moved during the study.}
    \label{fig:study_sites}
\end{figure}

\begin{figure}[t]
    \centering
    \includegraphics[width=0.75\textwidth]{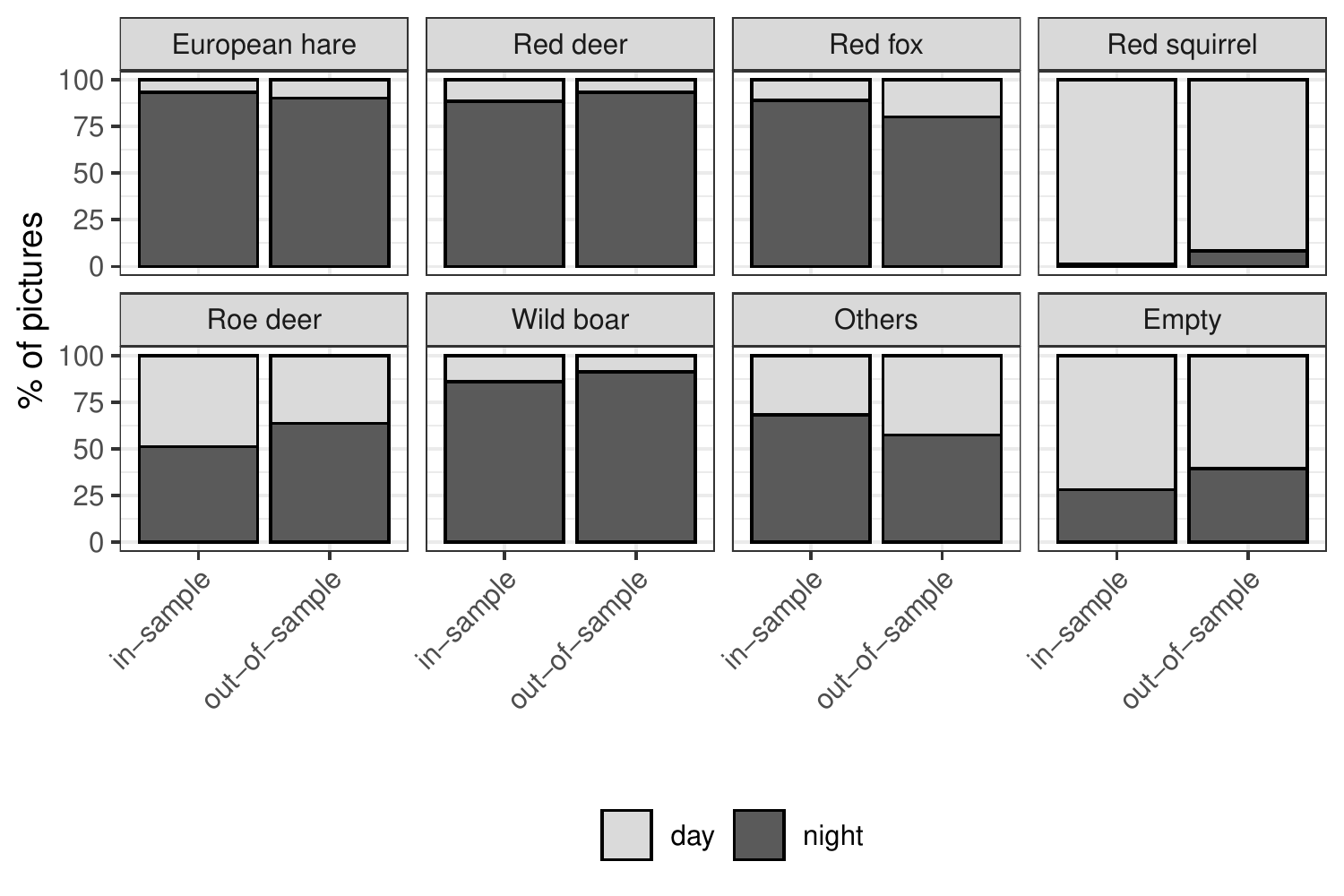}
    \caption{Distribution of day and night images per class for in-sample and out-of-sample images.}
    \label{fig:day_night}
\end{figure}

We used camera trap images from 37 wildlife camera traps installed in forests across the state of Bavaria in southeastern Germany (see Fig.~\ref{fig:study_sites}). 
The data set was collected along a gradient of human and climatic variation as part of a comprehensive study focusing on biodiversity along these gradients \citep{redlich_disentangling_2022}. 
The images used in this study are from the time period September 2019 to February 2020 (autumn-winter). All images containing humans, vehicles, and domesticated animals were removed in compliance with privacy laws. The total number of 48,116 images was divided into in-sample and out-of-sample data by camera stations, i.e., 18 camera stations along with their 24,368 images were declared as in-sample and the remaining 19 camera stations along with their 23,748 images were declared as out-of-sample. Table \ref{tab:train_data} summarizes the number of images per class for both data sets. Fig.~\ref{fig:day_night} shows ratios of day/night images. Note that for some classes, the ratios for in-sample and out-of-sample images are quite different, which reflects typical real-world situations. A desideratum of a good pipeline is to perform well also on these classes.

\begin{table}
\centering
\caption{Number of images per class.}
\begin{tabular}{lrrr}
\label{tab:train_data}
Species & in-sample & out-of-sample & total \\
\midrule
european hare & 485 & 255 & 740\\
red deer     &26 &   135 & 161\\
red fox      &708 &  106 & 814\\
red squirrel & 297 &  13 & 310\\
roe deer     & 6,074& 7,010 &13,084\\
wild boar  &210 & 904 &  1,114\\
others     &941 & 1,051&   1,992\\
empty       &15,627 & 14,274 &  29,901\\
\midrule
total & 24,368 & 23,748 & 48,116\\
\bottomrule
\end{tabular}
\end{table}

\subsection{Pipeline for tuning and model training}
\label{sec:methods-train-tune}

We consider a data set $\D = \Dset$ of labeled images, where $\yi \in \gset$ is the label of image $\xi$ and $g$ is the number of classes. 
The goal is to train a classification pipeline $\pihf$ that maps images $\xv$ to probability scores $\pixh$ of predicted hard labels $\yh = \argmax_{k \in \gset} \pikxh[k]$. Our pipeline consists of two major parts: In the first part, bounding boxes of animals in the images are detected via object detection. In the second part, bounding-box images are classified with respect to animal species.

\subsubsection{Classification pipeline with object detection and image classification} \label{sec:ppl}

\paragraph{Object detection -- from image level to bounding-box level} Often, a considerable fraction of each image does not show an animal (see Fig.~\ref{fig:misclass} for some examples). If the entire image is passed to an image classifier, the classifier needs to cope with the challenge of detecting the important pixels and focusing on those. 
Predictive performance can be increased by providing bounding boxes of animals. Different object detection pipelines have been proposed in the literature; we use the \textit{Megadetector} (MD) pipeline \citep{beery_efficient_2019} for its high usability and good performance in prior studies \citep[see][for a comparison]{velez_choosing_2022}. 
MD produces (among other results which we do not use in the following) for every image $\xi$ a (possibly empty) set of $m$ bounding-box coordinates 
with corresponding confidence values $\bm{c}^{(i)} =(c^{(i)}_1, \dots, c^{(i)}_m) \in [0,1]^m$ 
of capturing an animal. 
All bounding boxes with confidence $c^{(i)}_j \geq \alpha$ (\textit{high-conf BB}) are passed on to the next step -- the image classifier on the bounding-box level. All images that have no bounding boxes or only bounding boxes with a low confidence $c^{(i)}_j < \alpha$ (\textit{other BB}) are considered empty. 
The confidence threshold $\alpha$ is either set to a fixed value beforehand 
or can be chosen in a data-driven manner and specifically tailored for the data set at hand. 
We opt for the latter approach and consider the threshold $\alpha$ to be a hyperparameter optimized during tuning.

\paragraph{Image classification at the level of bounding boxes} We consider the following preprocessing on the level of bounding-box images (bb-images). 

\textit{(1) Cropping and resizing:} For all bounding boxes with a confidence threshold of at least $\alpha$, we crop the respective part from the original image and resize it, such that we obtain $s$ square bb-images $\bm{z}^{(i)}, i \in \{1, \dots, s\}$ of $224 \times 224$ pixels, which is the optimal input size for networks pre-trained on ImageNet \citep{deng_imagenet_2009}. 
We assign each bb-image the label of the original image that it is part of; this only makes sense if images are pure in the sense that if an image shows animals, it only shows animals of a unique species. In our data, we did not encounter any image with mixed animal species, 
but this assumption needs to be asserted critically for other data sets -- demanding to label on bounding-box level if the purity assumption does not hold.

\textit{(2) Augmentation:} We augment the bb-images of the training data by 
optional rotation, flipping, and changing the contrast, and use up to three randomly sampled augmentations per image for the analyses in this paper \footnote{
Technically speaking, augmentation is realized on-the-fly during training to avoid increased memory consumption.
}. 

\textit{(3) Transfer learning:} As a last preprocessing step, we carry out a forward pass through a pre-trained image classification network, choosing between the available architectures, where we exclude the final fully-connected layers.
The resulting data representation, whose structure the network has learned from millions of pre-training samples, encapsulates latent image properties from which classes can be predicted quite easily.

With the thus preprocessed data, we train an image classification model.

\paragraph{Predicting new images with the pipeline}
\label{sec:methods-predict}

Consider a set $\D_{new} = (\xi[1], \dots, \xi[w])$ of new images: 
First, we apply object detection, cropping and resizing, and the forward pass through the pre-trained model. 
The resulting data is used to predict the class scores 
of the bb-images $\D_{z,new}$ with the image classifier $\pixh$.
Then, the results of object detection and image classification are merged: The predicted label is set to ``empty'' for all images which (i) have no bounding boxes with a confidence of at least $\alpha$ or (ii) have bounding boxes that pass the threshold but are classified as ``empty'' by the image classifier.
For all other images, the predicted label is set to the non-empty class with the highest weighted average of the bounding box predictions, where the MD confidences are used as weights.
This weighted average is retained as final confidence about the class. 
This yields

\begin{enumerate}[label=(\roman*)]
    \item predicted labels $\yih$ for each image $\xi[i] \in \D_{new}$,
    \item average weighted predicted class scores $\hat{\pi}_k^{(i)}$ for each class $k \in \gset$ per image, 
    \item the total predicted number $o_k^{(i)}$ of each animal class per image, and
    \item a final confidence score $d^{(i)} = \max_{k \in \gset} \hat{\pi}_k^{(i)}$ per image, reflecting the highest class-wise score.
\end{enumerate}

Optionally, images whose highest confidence for any class  
does not exceed a user-specified threshold, i.e., $d^{(i)} \leq \beta, $ may still remain ``unlabeled'', thus giving the user the opportunity to label it manually in a later iteration of the AL process.

\subsubsection{Tuning pipeline} 
\label{sec:tuning}

\paragraph{Resampling strategy}

\begin{figure}[t]
    \centering
    \includegraphics[width=14.4cm]{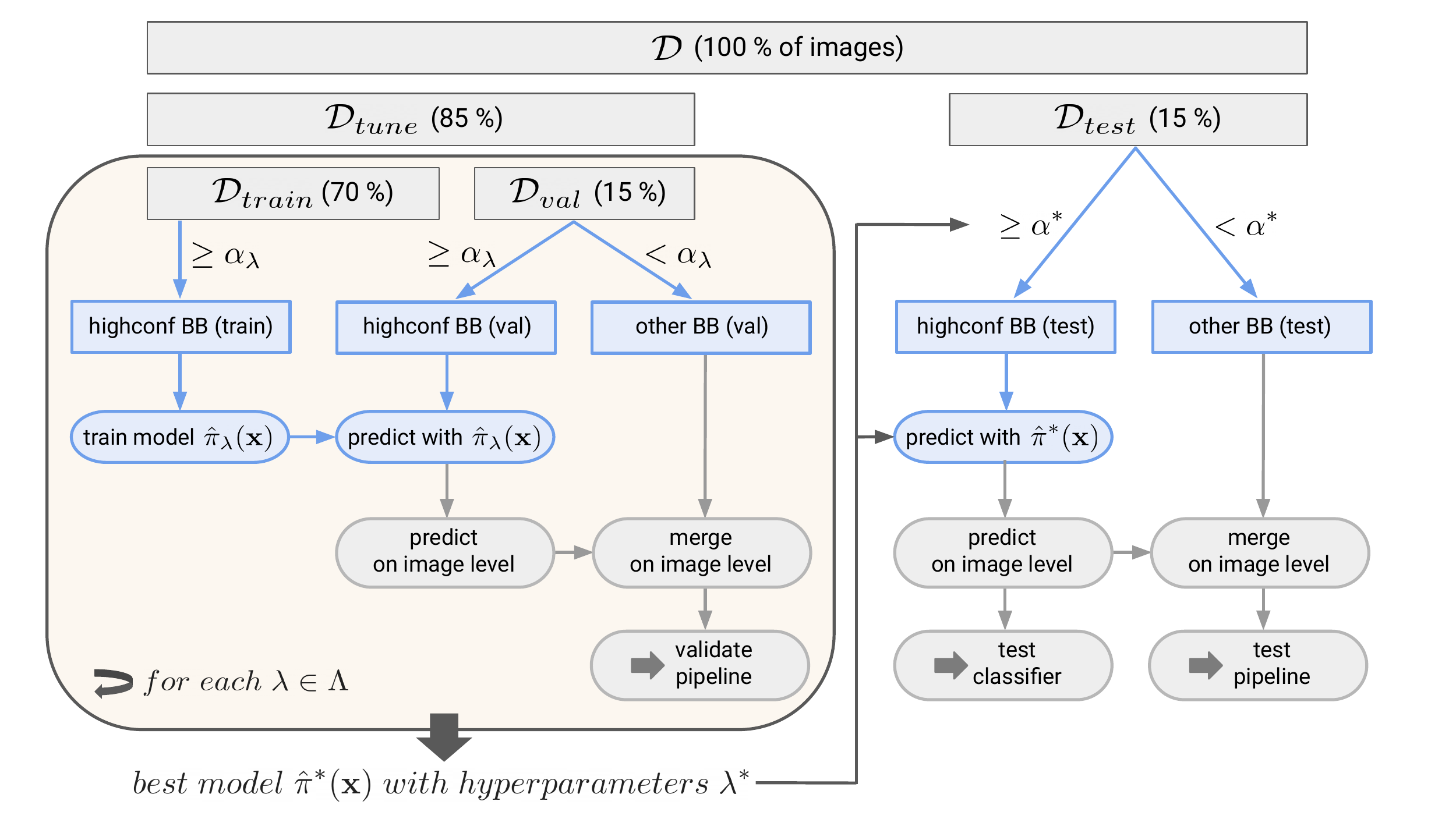}
    \caption{Resampling scheme. Hyperparameter tuning is carried out on tuning data $\Dtune$. Generalization errors using the best hyperparameters are estimated on untouched test data $\DTest$.}
    \label{fig:resampling}
\end{figure}

The above pipeline comprises hyperparameters (\begin{enumerate*}[label=(\roman*)]
    \item megadetector threshold
    \item choice of the pre-trained network
\end{enumerate*}) which cannot be learned during training but need to be optimized via hyperparameter tuning. 
Fig.~\ref{fig:resampling} visualizes the resampling strategy and pipeline for tuning and evaluation:
We divide $\D$ randomly (stratified by camera station) into data for tuning and training ($\Dtune$) and data for evaluating the model performance ($\DTest$), where $15\%$ are used for $\DTest$. Tuning data $\Dtune$ is further split into training data $\DTrain$ and validation data $\Dval$, such that those sets comprise $70\%$ and $15\%$ of $\D$, respectively. Training data $\DTrain$ is then used to train a classification model on the level of bounding boxes for each combination of hyperparameters; the best hyperparameter combination is chosen with respect to the performance of the pipeline on validation data $\Dval$. Finally, generalization errors of the classifier and the pipeline, endowed with the chosen hyperparameters, are estimated on the untouched test data $\DTest$.

\begin{tune}
\paragraph{Tuning}
We train a model $\pih_{\lamv}(\cdot)$ for each hyperparameter combination $\lamv \in \Lam$ -- where $\Lam$ is the corresponding search space -- on $\D_{z,train}$ and evaluate the pipeline of object detection and image classification with this model $\pih_{\lamv}(\cdot)$ on $\Dval$. 

Predicted labels are computed by combining the results of object detection and image classification -- as explained above for new images.
Evaluation metrics $\rho_{\lamv}$ are computed 
for each hyperparameter combination separately. 
For our tuning process, one of the four alternatives recall, precision, F1-score, or accuracy can be chosen by the user -- for the results presented in this paper, we choose the weighted F1-score.
The hyperparameter combination yielding the highest $\rho_{\lamv}$ on $\Dval$, i.e., 
$\bm{\lambda}^* = \argmax_{\lamv \in \Lam} \rho_{\lamv}$, is considered the best hyperparameter combination, the corresponding model is 
$\pi^*(\cdot)$.

\paragraph{Evaluation at the level of bounding boxes}  
We first carry out the object detection and the above preprocessing steps of cropping, resizing, and the forward-pass through the pre-trained model -- as indicated by the respective value in $\bm{\lambda}^*$ -- on $\DTest$. The labels of the resulting bb-images are predicted by $\pi^*(\xv)$ and an evaluation metric can be computed for the high-conf bb-images in $\D_{z,test}$. 

\paragraph{Evaluation at the level of original images} As above, we merge the results of object detection and image classification for the images in $\DTest$ and compute the respective evaluation metric.

\end{tune}

\subsection{Pipeline for active learning}
\label{sec:methods-active}

We propose the following active learning (AL) pipeline (see also Algorithm \ref{alg:al}).

\textit{(1) Initialization:} A data set of unlabeled images is provided by the user. If for some images, labels are already available, these can be used and hence no random sampling is necessary in step (2a). However, since the selection of images that are labeled might not be representative of the entire data set in practice, e.g., because all labels are from a very small and hence similar period of time, we generally recommend random sampling in the first iteration. If available, information about which images belong to the same camera station should be provided, too, to allow for stratification in the resampling process.

\textit{(2) AL loop:} As long as the user is unsatisfied with the prediction performance of the current model (as reported by the evaluation step -- and optionally by visual inspection), the active learning loop is carried out: 

\textit{(2a) Image selection:} A user-specified number of unlabeled images is selected by the AL engine. 
These are the images that have the highest value of a so-called ``acquisition function''; the value of this function reflects the potential of an image for improving the model -- we use softmax entropy. 
Optionally, this can be done stratified by camera stations. 
In the first iteration of the AL loop, where no model is available, image selection is done randomly.

\textit{(2b) Manual labeling:} The selected images are labeled by a human expert.

\textit{(2c) Model training:} A full step of tuning, training, and evaluation as described above is carried out, based on the available labeled data. 
If the time budget is limited, the user may consider skipping tuning here.

\textit{(3) Predict remaining unlabeled images:} As the last step of the pipeline, all remaining unlabeled images are predicted by the final model.

For the evaluation of the trained models, a sufficiently large subset of the labeled images needs to be set apart as a test set. This test set is used only for evaluating the performance of the model and remains unchanged during the AL loop. 

\begin{algorithm}
\caption{Active Learning Workflow (H: Human -- M: Machine)}
\label{alg:al}
\begin{algorithmic}
\State H: Load data (unlabeled or partly labeled) \Comment{\textit{(1)}}
\If{No labeled images} 
    \State M: Select random images to be labeled \Comment{\textit{(2a)}}
    \State H: Label images manually \Comment{\textit{(2b)}}
\EndIf
\State M: Tune, train, evaluate initial model \Comment{\textit{(2c)}}
\While{H: Not satisfied with test performance} \Comment{\textit{(2)}}
    \State M: Select best images to be labeled via acquisition function \Comment{\textit{(2a)}}
    \State H: Label images manually \Comment{\textit{(2b)}}
    \State M: Tune, train, evaluate \Comment{\textit{(2c)}}
\EndWhile
\State M: Predict all remaining unlabeled images with the last model \Comment{\textit{(3)}}
\end{algorithmic}
\end{algorithm}

\section{Results}

In the following, we show results for the Bavarian data presented in Section \ref{sec:data}, Appendix \ref{sec:app} applies our method to a different dataset, obtained from the LILA database.

\subsection{In-sample results}
\label{sec:res_ins}

For tuning, training, and evaluating the models, we use a $70\%$-$15\%$-$15\%$ train-val-test split and report evaluation metrics only for the test set which is not seen during training or tuning of the models. As data, we use the 24,368 ``in-sample'' images shown in Table \ref{tab:train_data}. 
As tuning search space $\Lam$ we use a grid consisting of all combinations of architecture $\in$ \{InceptionResNetV2, Xception, DenseNet121\}, which are all state-of-the-art across many image classification tasks and parameter-efficient at the same time, and MD threshold $\alpha \in \{0.1, 0.3, 0.5, 0.7, 0.9\}$. 
When writing this manuscript, the current version of MegaDetector is MDv5, which we use for our analyses (from the two versions MDv5a and MDv5b, we choose the former which is trained on more data). However, we also compare tuning results using the prior version MDv4 for two reasons: (a) Some of the seminal papers in wildlife camera trap image classification such as \cite{norouzzadeh_deep_2021} use MDv4 with $\alpha=0.9$ without tuning and we want to make a sensible comparison here, and (b) the optimal confidence threshold may have changed in comparison to older versions, which we want to investigate for a fixed dataset, ruling out differences in the images analyzed as a reason for differing results.\footnote{The authors of MD give a similar word of caution on choosing confidence thresholds in their release note of MDv5 -- without hinting on how to optimize this properly: \url{https://github.com/microsoft/CameraTraps/releases}}
All reported numbers represent the average over results from three independent runs.
        
\subsubsection{Benefit of tuning hyperparameters}
\label{sec:tuning_results}

Table \ref{tab:tuningmdv5} shows the weighted F1-score on $\Dval$ for the 8-class classification for different choices of the hyperparameters using MDv5. 
As can be seen, tuning the hyperparameters properly has a substantial impact on the predictive performance of the model. 
The bold first row reflects the optimal choice of hyperparameters which is used for the following experiments.
Table \ref{tab:tuningmdv4} shows results for MDv4. We observe two key insights: (a) our pipeline with MDv5 is performing substantially better than with MDv4, and (b) the optimal hyperparameter configuration is differing very much. The second point underlines the importance of tuning the hyperparameters since there is no guarantee that for (a) other datasets or (b) other (versions of) object detectors the optimal hyperparameters are the same. Interestingly, the performance is more variable for MDv5 than for MDv4, meaning that proper tuning has more impact for the newer version. 

\begin{table}[h]
\centering
\caption{Hyperparameter tuning: best and worst configurations sorted by F1-score (MDv5)}
\begin{tabular}{lrr}
\label{tab:tuningmdv5}
Confidence & Architecture & F1-score\\
\midrule
\textbf{0.1} & \textbf{Xception} & \textbf{0.933} \\
0.3 & Xception & 0.932 \\
0.5 & Xception & 0.921 \\
0.7 & Xception & 0.900 \\
$\vdots$ & $\vdots$ & $\vdots$ \\
0.9 & Xception & 0.781 \\
0.9 & DenseNet121 & 0.780 \\
0.9 & InceptionResNetV2 & 0.777 \\
\bottomrule
\end{tabular}
\end{table}

\begin{table}[h]
\centering
\caption{Hyperparameter tuning: best and worst configurations sorted by F1-score (MDv4)}
\begin{tabular}{lrr}
\label{tab:tuningmdv4}
Confidence & Architecture & F1-score\\
\midrule
\textbf{0.5} & \textbf{Xception} & \textbf{0.909} \\
0.7 & Xception & 0.908 \\
0.9 & InceptionResNetV2 & 0.908 \\
$\vdots$ & $\vdots$ & $\vdots$ \\
0.3 & DenseNet121 & 0.899 \\
0.1 & InceptionResNetV2 & 0.894 \\
0.1 & DenseNet121 & 0.888 \\
\bottomrule
\end{tabular}
\end{table}

\subsubsection{Empty vs. non-empty images}

During tuning, the best confidence threshold was 
$\alpha=0.1$ for MDv5 and $\alpha=0.5$ for MDv4, respectively. 
We can now compare results for the binary classification of ``empty'' vs. ``non-empty'' for different choices of $\alpha$ in Table \ref{tab:empty_resultsmdv5} for MDv5 and Table \ref{tab:empty_resultsmdv4} for MDv4.
The empty recall increases with $\alpha$ due to the fact that the MD is considering more images as empty.
On the other hand, the non-empty recall decreases with $\alpha$, meaning that more non-empty images (i.e., animals) are retrieved for smaller $\alpha$. 
In fact, with tuned $\alpha=0.5$ for MDv4, the false-empty rate is $14.2\%$ less than with $\alpha=0.9$ (a common choice, e.g., \cite{norouzzadeh_deep_2021}) ($10.9\%$ vs. $12.7\%$), meaning that $14.2\%$ fewer animal images are overlooked.\footnote{This effect is much stronger for MDv5 but \cite{norouzzadeh_deep_2021} used MDv4, which is why we compare numbers for MDv4 here.} From a practical perspective, this is highly desirable: While it is no big effort to manually discard some images that are empty but labeled as non-empty, it is not feasible to detect non-empty images that are labeled as empty without screening all the images again -- which is exactly what we wish
to avoid by using an automatic image classifier.
This shows that tuning $\alpha$ is also beneficial for the ``empty'' vs. ``non-empty'' performance and increases the number of detected animals overall.

\begin{table}[h]
\centering
\caption{Empty vs. non-empty images (MDv5), ``no MD'' shows metrics for not using object detection at all, hence only image classification on entire images.}
\begin{tabular}{l|c|ccc|ccc}
\label{tab:empty_resultsmdv5}
& & & empty & & & non-empty & \\
\hline
Confidence & Accuracy & Precision & Recall & F1 & Precision & Recall & F1 \\[.1ex]
\hline
no MD  &  0.920 & 0.913 & 0.969 & 0.940 & 0.936 & 0.830 & 0.880\\
0.1 & 0.962 & 0.950 & 0.993 & 0.971 & 0.986 & 0.906 & 0.944 \\
0.3 & 0.959 & 0.946 & 0.994 & 0.969 & 0.987 & 0.899 & 0.941 \\
0.5 & 0.947 & 0.927 & 0.996 & 0.960 & 0.991 & 0.861 & 0.922 \\
0.7 & 0.927 & 0.898 & 0.998 & 0.946 & 0.996 & 0.799 & 0.887 \\
0.9 & 0.816 & 0.777 & 0.999 & 0.874 & 0.997 & 0.491 & 0.658 \\
\bottomrule
\end{tabular}
\end{table}

\begin{table}[h]
\centering
\caption{Empty vs. non-empty images (MDv4).}
\begin{tabular}{l|c|ccc|ccc}
\label{tab:empty_resultsmdv4}
& & & empty & & & non-empty & \\
\hline
Confidence & Accuracy & Precision & Recall & F1 & Precision & Recall & F1 \\[.1ex]
\hline
0.1  &  0.949 & 0.950 & 0.973 & 0.961 & 0.948 & 0.907 & 0.927\\
0.3  &  0.952 & 0.947 & 0.981 & 0.964 & 0.963 & 0.900 & 0.930\\
0.5 & 0.949 & 0.943 & 0.982 & 0.962 & 0.964 & 0.891 & 0.926 \\
0.7 &  0.952 & 0.940 & 0.990 & 0.964 & 0.979 & 0.885 & 0.930\\
0.9 &  0.950 & 0.934 & 0.991 & 0.962 & 0.981 & 0.873 & 0.924\\
\bottomrule
\end{tabular}
\end{table}

\subsubsection{Multi-class model performance}

\begin{figure}[h]
\centering
\begin{fig_publish}
\includegraphics[width=.7\linewidth]{fig/cm_insamplemd5.eps}
\end{fig_publish}
\begin{fig_count_words}
\includegraphics[width=.7\linewidth]{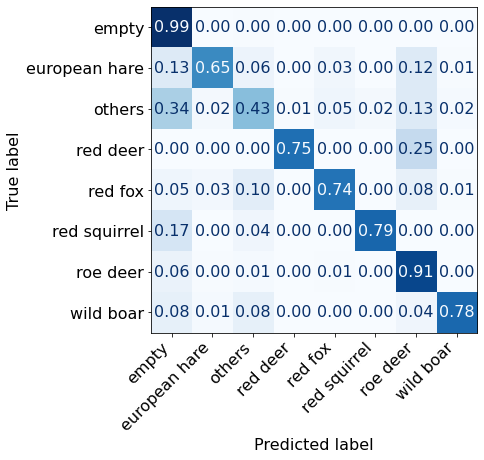}
\end{fig_count_words}
\caption{In-sample performance}
\label{fig:multi_conf}
\end{figure}

From here on, we are focusing on the results of MDv5. Table \ref{tab:ins-oos}, first and second row, lists overall metrics of the classification performance on $\DTest$. The first row contains results of omitting object detection and directly performing image classification on the entire images; a comparison with the second row reveals a clear benefit of using the pipeline of object detection and image classification.
Fig.~\ref{fig:multi_conf} shows the detailed predictive performance of the tuned pipeline.  
Performance for larger classes (in terms of the number of images, not of animal size) such as ``roe deer'' is better than for smaller classes such as ``wild boar''.
The class ``other'' proves difficult to detect, which might be due to the fact that this class is rather heterogeneous as it consists of (i) other animal species than considered here (hence low numbers per class) and (ii) images where the true animal class could not be determined by the human annotators with sufficient confidence. 

\begin{table}[h]
\centering
\caption{Multi-class classification (MDv5). Precision and F1-score are weighted averages of 8 classes, (weighted recall is not shown since equal to accuracy by definition).}
\begin{tabular}{lrrrr}
\label{tab:ins-oos}
{} & Accuracy & Precision &  F1\\
\midrule
no MD (in-sample) &  0.865 & 0.869 &  0.856 \\
In-sample &  0.933 & 0.929 &  0.930 \\
Out-of-sample & 0.868 & 0.874  & 0.867 \\
Active learning -- 37$\%$ & 0.926 & 0.921 &  0.919 \\
Active learning -- 100$\%$& 0.932 & 0.929 &  0.927 \\
\bottomrule
\end{tabular}
\end{table}

\begin{figure}
     \centering
     \begin{subfigure}[b]{0.48\textwidth}
         \centering
         \includegraphics[width=\textwidth]{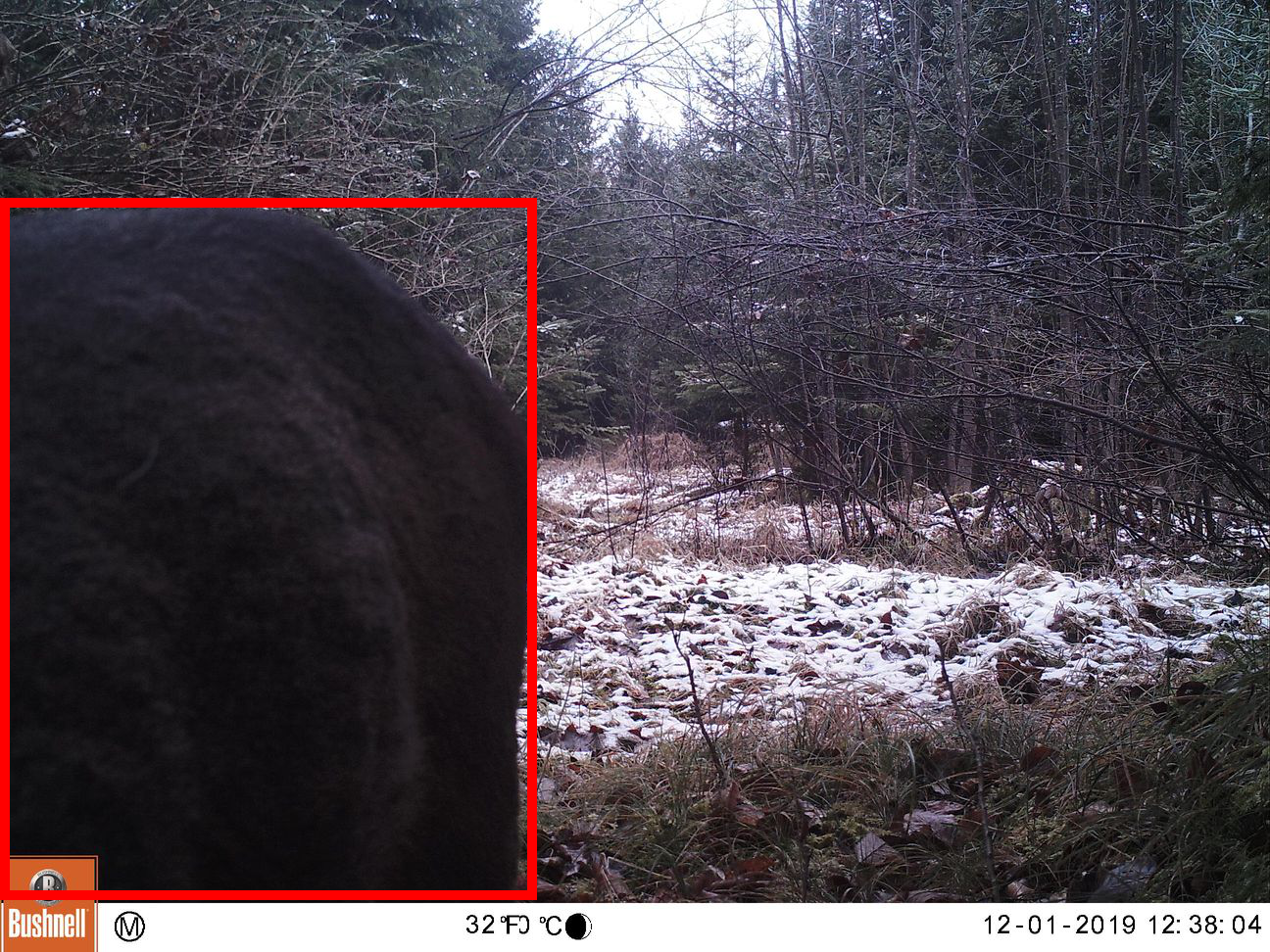}
         \caption{\centering Extreme close-up. \textit{Label}:~roe deer, \textit{predicted}: others, \textit{confidence}: 0.74
         \vspace{0.5cm}
         }
         \label{fig:wrong_close}
     \end{subfigure}
     \hfill
     \begin{subfigure}[b]{0.48\textwidth}
         \centering
         \includegraphics[width=\textwidth]{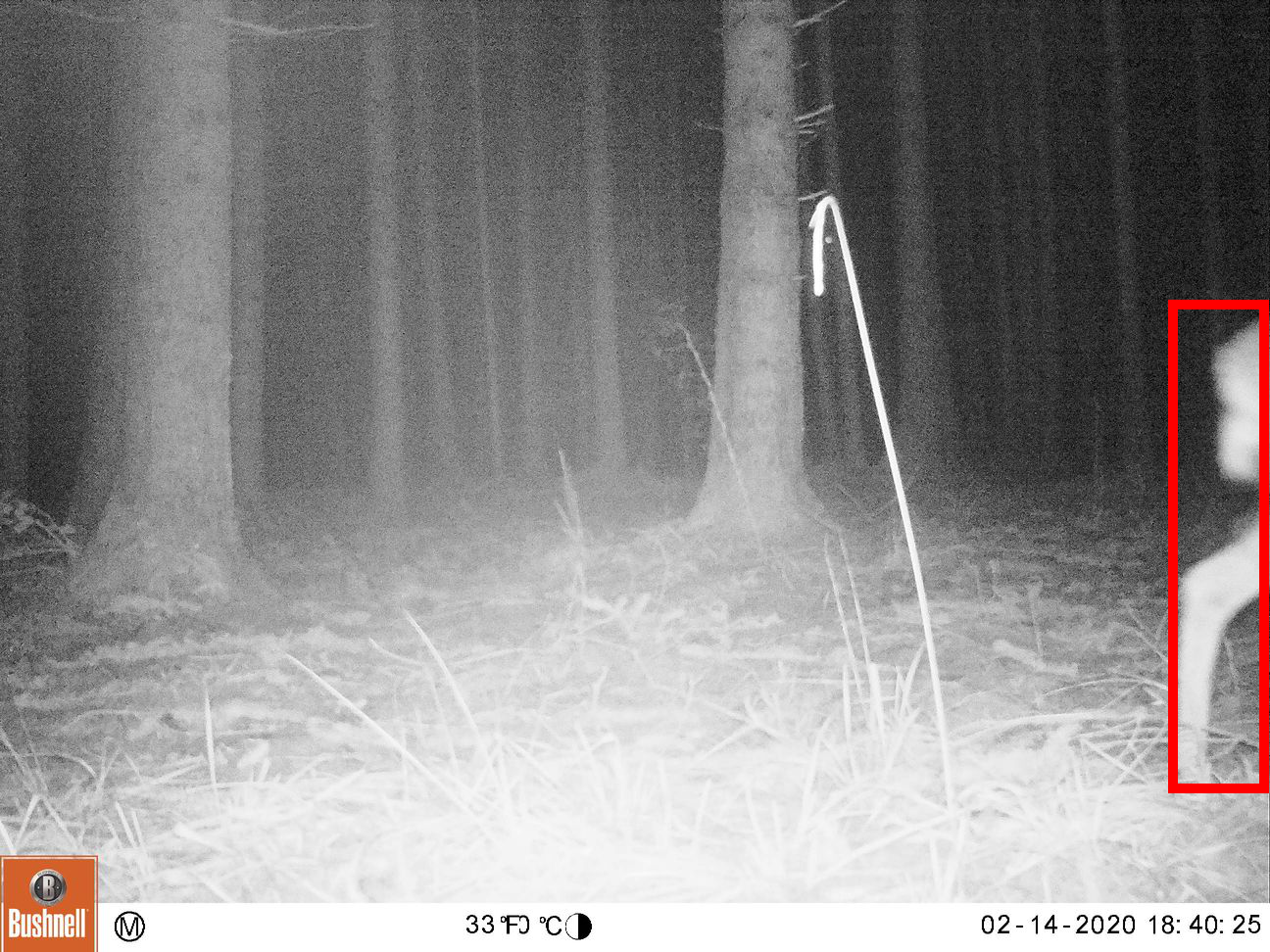}
         \caption{\centering Partial occlusion. \textit{Label}:~roe deer, \textit{predicted}: others, \textit{confidence}: 0.67
         \vspace{0.5cm}
         }
         \label{fig:occlusion}
     \end{subfigure}
     \hfill
          \begin{subfigure}[b]{0.48\textwidth}
         \centering
         \includegraphics[width=\textwidth]{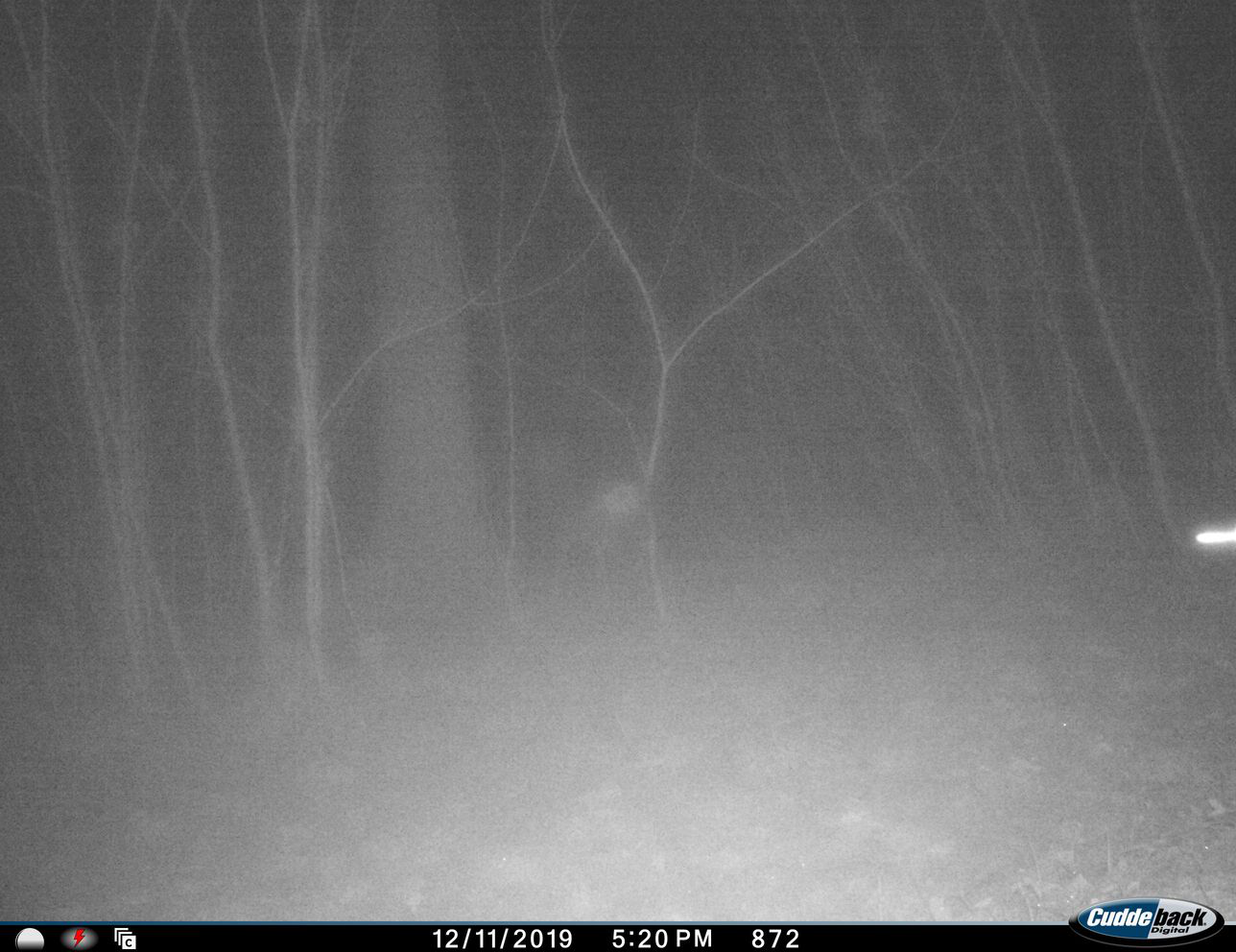}
         \caption{\centering Poor light. \textit{Label}: roe deer, \textit{predicted}:~empty, \textit{confidence}: 1.00
         \vspace{0.5cm}
         }
         \label{fig:bad_light}
     \end{subfigure}
     \hfill
     \begin{subfigure}[b]{0.48\textwidth}
         \centering
         \includegraphics[width=\textwidth]{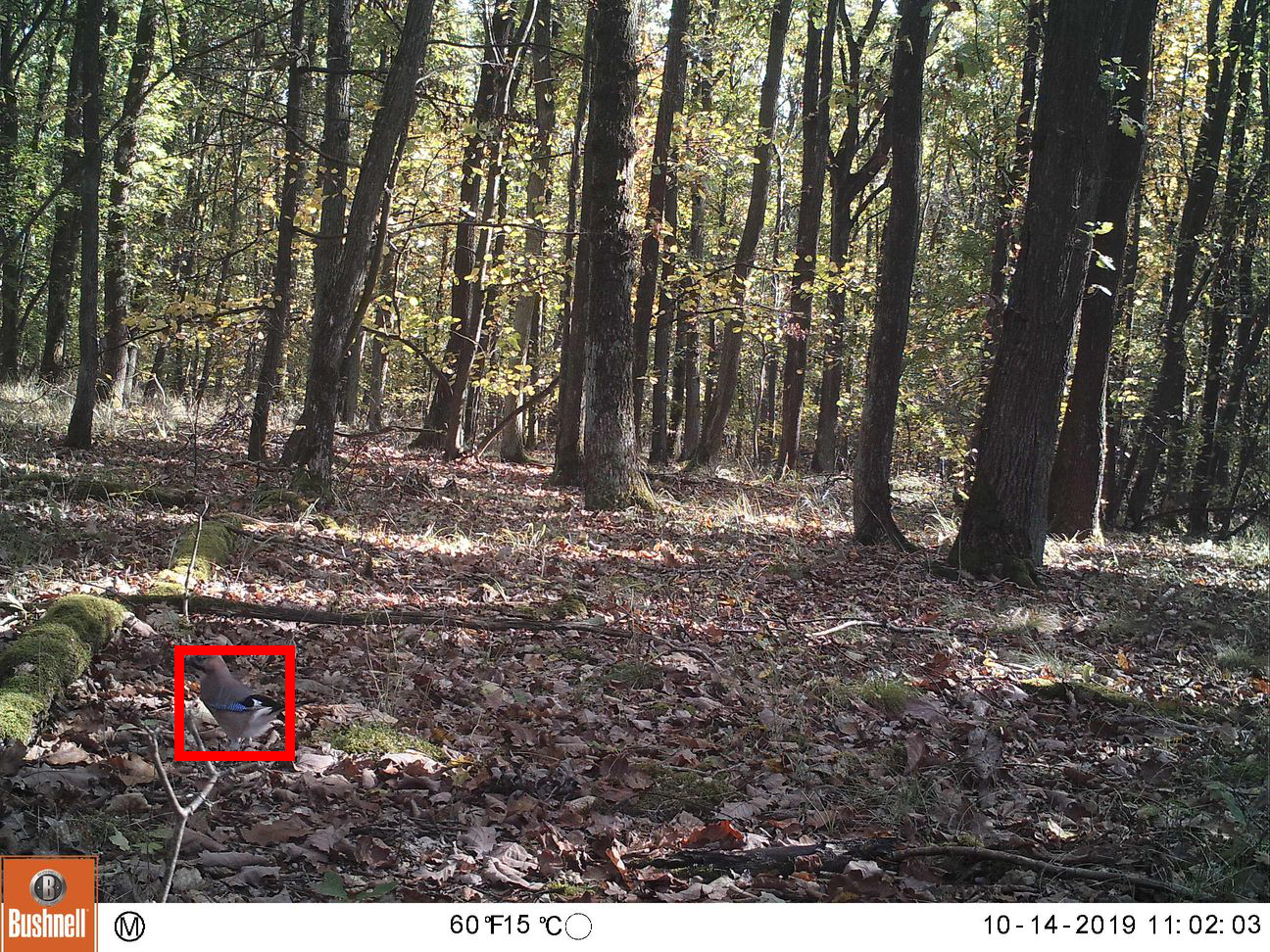}
         \caption{\centering Small objects. \textit{Label}: others, \textit{predicted}:~red squirrel, \textit{confidence}: 0.43
         \vspace{0.5cm}
         }
         \label{fig:wrong_tiny}
     \end{subfigure}
     \hfill
    \begin{subfigure}[b]{0.48\textwidth}
         \centering
         \includegraphics[width=\textwidth]{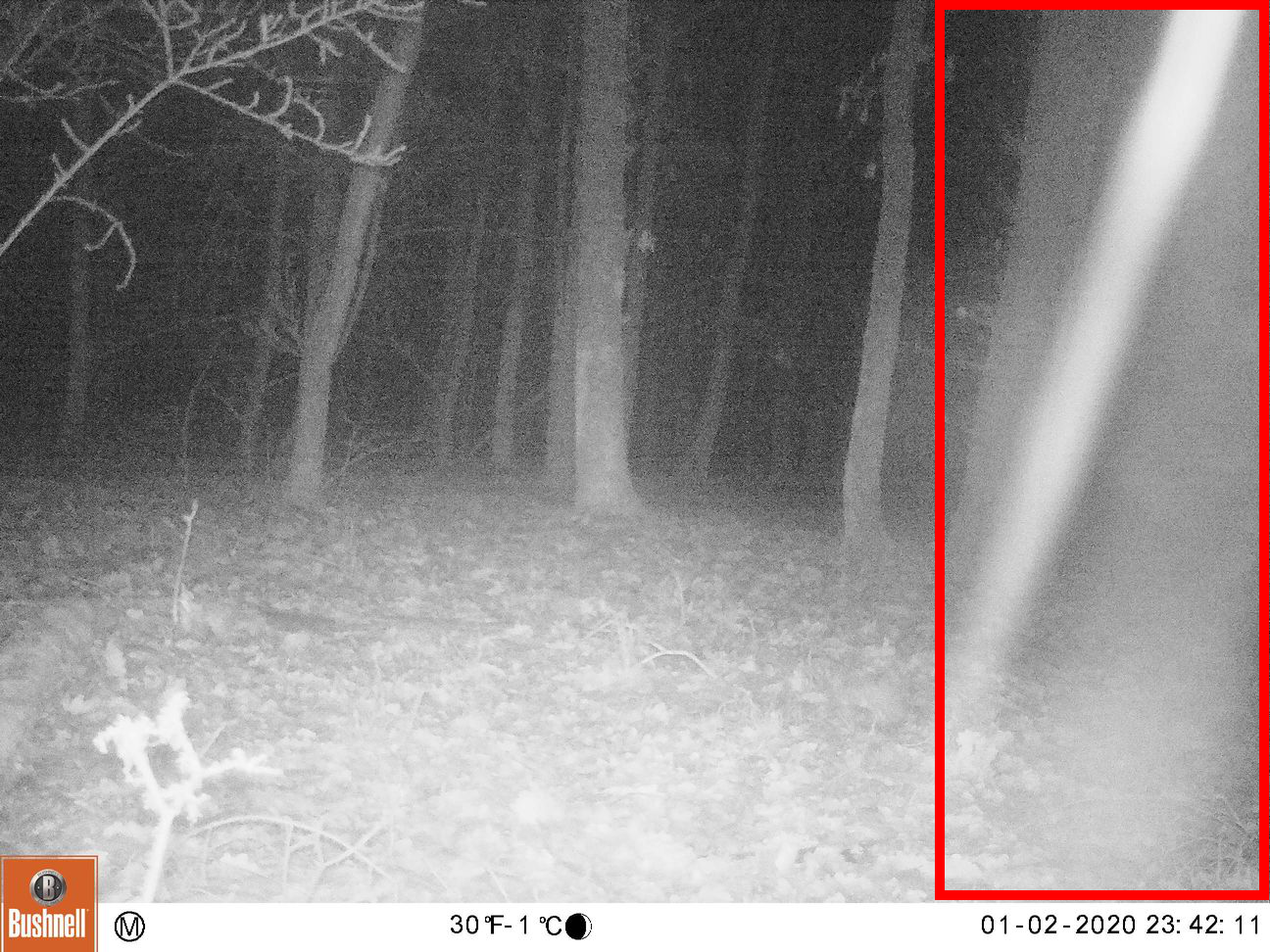}
         \caption{\centering Wrong bounding box. \textit{Label}: empty, \textit{predicted}:~roe deer, \textit{confidence}: 0.65
         \vspace{0.5cm}
         }
         \label{fig:wrong_bb}
     \end{subfigure}
     \hfill
     \begin{subfigure}[b]{0.48\textwidth}
         \centering
         \includegraphics[width=\textwidth]{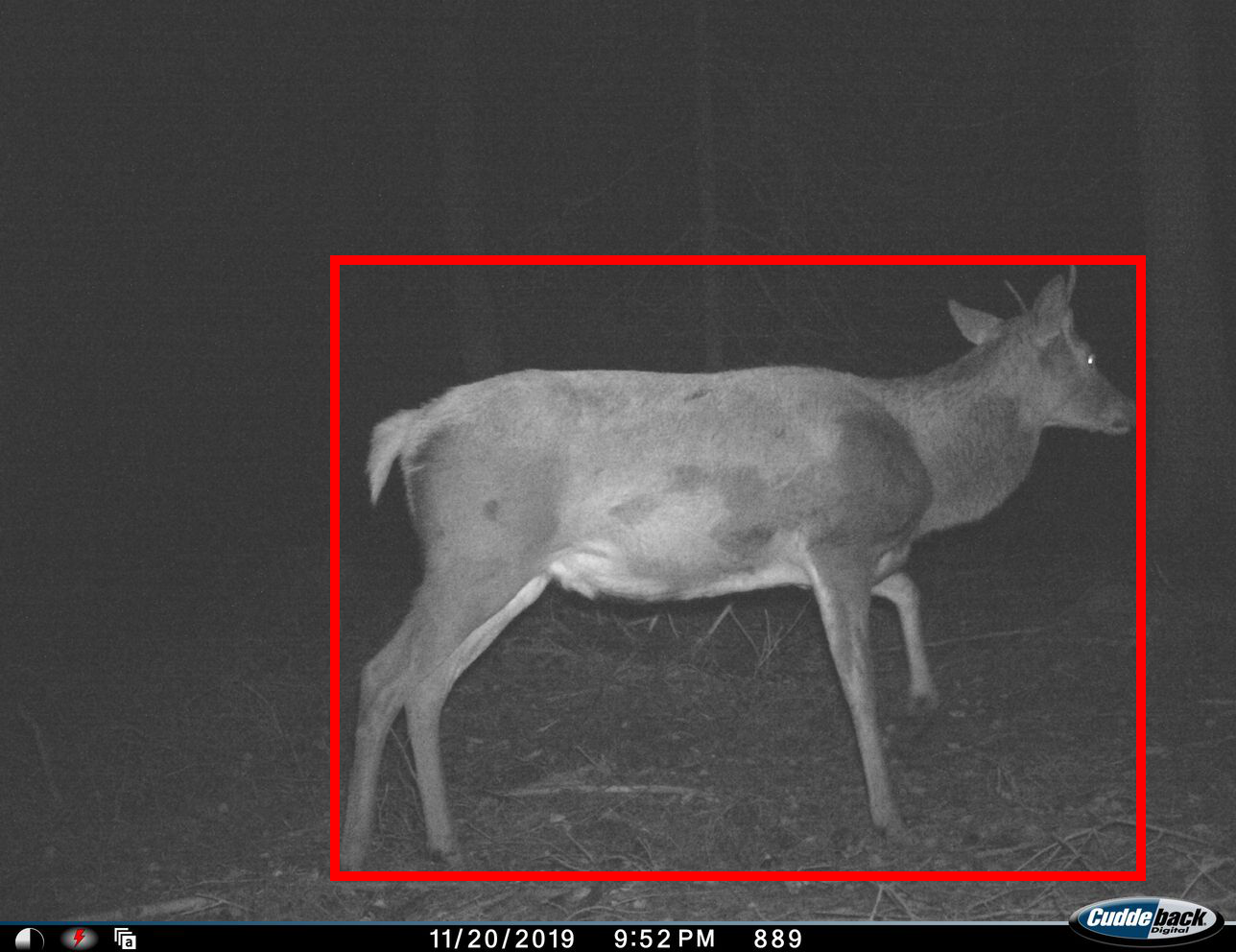}
         \caption{\centering Wrong human label. \textit{Label}: others, \textit{predicted}:~roe deer, \textit{confidence}: 0.65
         \vspace{0.5cm}
         }
         \label{fig:wrong_label}
     \end{subfigure}
    \caption{Misclassified images}
    \label{fig:misclass}
\end{figure}

Fig.~\ref{fig:misclass} displays some misclassified images and points to cases where the classification is wrong. Reasons seem to be, among others: Extreme close-up (\ref{fig:wrong_close}), difficulties due to partial occlusion (\ref{fig:occlusion}), poor lighting -- and hence no bounding box by MD (\ref{fig:bad_light}), very small objects (\ref{fig:wrong_tiny}), wrong bounding box by MD (\ref{fig:wrong_bb}), and wrong human label (\ref{fig:wrong_label}). In the last case, the prediction counts as misclassified since the human label and the predicted label do not match, but note that actually both the prediction and the human label are wrong in this case, since this is a red deer, not a roe deer.

\subsection{Out-of-sample results} \label{sec:oosample}

For a sensible comparison of OOS results, i.e., comparing model performance without AL and with AL -- and also different stages of AL--, we sampled a random test set from the OOS images, consisting of 15\% of the images, stratified by camera station. This test set comprises 3,667 images, from which 1,503 images were identified as non-empty by the MD at confidence $\alpha = 0.1$, is never used for training or tuning, and remains unchanged over all model comparisons.

\subsubsection{Model performance out-of-sample without active learning}

\begin{figure}
\centering
\begin{fig_publish}
\includegraphics[width=.7\linewidth]{fig/cm_oosamplemd5.eps}
\end{fig_publish}
\begin{fig_count_words}
\includegraphics[width=.7\linewidth]{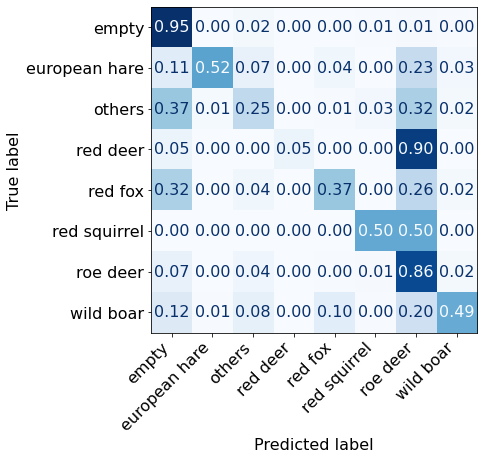}
\end{fig_count_words}
\caption{Out-of-sample performance without AL.}
\label{fig:oos_multi}
\end{figure}

Table \ref{tab:ins-oos}, third row, shows the predictive performance of predicting the OOS test data with the model trained on the entire in-sample data; the drop in predictive performance is expected and in line with results in the literature. Fig.~\ref{fig:oos_multi} shows detailed results of the image classifier for the OOS test data. A comparison with the class distribution between in-sample and out-of-sample data reveals that performance is especially low where the distribution has shifted most. For example, there are only 26 red deers in the entire training data; it is hence unsurprising that the predictive performance for this class will be rather low. This distribution shift between train and test data is exactly the problem of OOS prediction which shall be remedied by using active learning: By adding more and more training data from the new distribution, the performance should increase until a satisfying level is reached.

\subsubsection{Model performance out-of-sample with active learning}

\begin{figure}
\centering
\begin{fig_publish}
\includegraphics[width=\linewidth]{fig/active_learningmd5.eps}
\end{fig_publish}
\begin{fig_count_words}
\includegraphics[width=\linewidth]{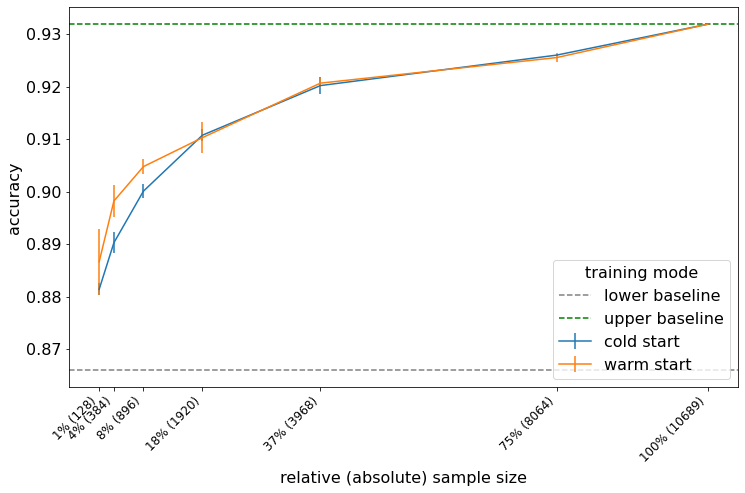}
\end{fig_count_words}

\caption{Active-learning performance (8-class accuracy) for cold and warm start as compared to out-of-sample prediction (lower horizontal line) and training on the entire data set (upper horizontal line). Error bars indicate variability over 3 different random seeds.}
\label{fig:oos_active}
\end{figure}

Fig.~\ref{fig:oos_active} shows the results of the active learning pipeline.
For a proper assessment of the benefit of active learning, we compare it with two baselines: The lower baseline is the model trained on the in-sample data (Table \ref{tab:ins-oos}, third row). The upper baseline is a model trained on the entire OOS data (Table \ref{tab:ins-oos}, last row). We compare two active learning strategies; for both strategies, we use the architecture that was found to be optimal in the above tuning procedure, i.e., the Xception architecture. 
The classification head (i.e., the set of final linear layers) is replaced by a new classification head with freshly initialized weights.
Note that this enables tailoring the pipeline to tasks with an arbitrary set of classes by simply adjusting the number of output units in the last layer (in contrast to directly using the model trained on the in-sample data which is consequently restricted to the classes seen during training).
The two strategies differ in the following:

\begin{itemize}
    \item \textbf{Cold-start:} For the rest of the neural net (the so-called backbone responsible for the task of learning a suitable data representation), the weights pre-trained on ImageNet are used.
    \item \textbf{Warm-start:} For the backbone, the weights pre-trained on ImageNet and on the in-sample data are used.
\end{itemize}

\begin{figure}
\centering
\begin{fig_publish}
\includegraphics[width=.7\linewidth]{fig/cm_activemd5.eps}
\end{fig_publish}
\begin{fig_count_words}
\includegraphics[width=.7\linewidth]{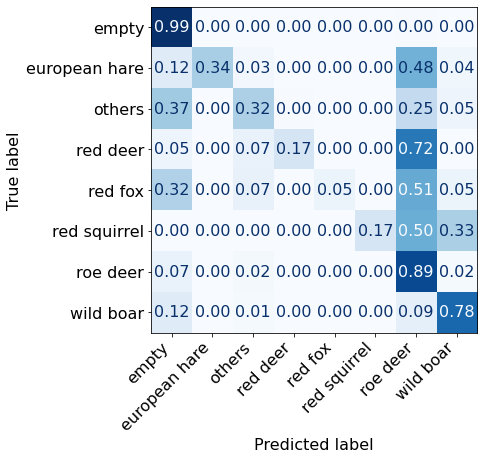}
\end{fig_count_words}
\caption{Out-of-sample performance with AL using $37\%$ of available data, i.e., 3,968 images.}
\label{fig:al_multi}
\end{figure}

As can be seen in the very left part of Fig.~\ref{fig:oos_active}, both active learning strategies outperform the lower baseline (which has never encountered samples from the OOS data) already in the initial iterations. This is somewhat surprising since in the first iteration only 128 images are used for training -- where the lower baseline is trained on the entire in-sample data of 24,368 images -- and underlines the importance of tailoring the model to the specific dataset at hand. 
With an increasing amount of images, the active learning models improve further. 
In the first iterations, warm-starting has a substantial advantage over cold-starting, this effect vanishes after using 1920 images for training. 
This means that the information in the OOS data is eventually better than the information in the in-sample data.
Using a relative sample size of $37\%$, i.e. 3,968 images, we already reach $99.4\%$ of the performance of the upper baseline, showing the benefit of using active learning.

Note the general applicability of the active learning approaches: Since we replace the classification head, the strategy can be used to learn models predicting any set of classes. This allows learning well-performing models with just a small number of manually annotated images. A further advantage is that the number of images to be labeled by a human does not have to be fixed beforehand: The active learning procedure can be carried out iteratively until a satisfying performance is reached, thus using the human annotation labor very efficiently. 
Furthermore, if the new images are expected to be rather similar to the existing ones (e.g., by setting up a new camera station in a territory comprising the same fauna), warm-starting may be beneficial and can decrease the number of images needed even more.

\section{Discussion and Conclusions}

We presented two methodological advances in using deep learning methods for wildlife image classification. First, a thorough tuning procedure for optimizing the hyperparameters of a multi-step pipeline, consisting of object detection and image classification. Second, an active learning component that enables efficient training of a high-performing model on new data, potentially from a new monitoring location or involving previously unseen animal species. We accompany the methodological developments with ready-to-use software which does not require programming skills of the user. Thereby, we leverage the potential of deep learning and active learning for a broad target group including all researchers in ecology, even with different analytical backgrounds.

Our results showed that tuning the hyperparameters has indeed an effect on the predictive performance of the resulting models. While taking well-performing hyperparameter choices from the literature might be a solid start in a project, tuning increases performance. The need for thorough tuning is even higher if no such hyperparameters choices from the literature exist, e.g., for new versions of an object detector like MegaDetector. 
The overall performance of our models is not competitive with some contributions in the literature which is due to the fact that we aim at smaller projects and consider a data set of around 50,000 images, while others use several millions of images. However, we show that the active learning component quickly helps improve the results for out-of-sample images and that warm-starting additionally increases performance in the first iterations.
Note that while we carefully designed in-sample and out-of-sample data sets making sure that no image of the out-of-sample set stems from a camera station contained in the in-sample set, these images are still from comparable regions in Bavaria, Germany. Transferring this to other regions of the world and hence to backgrounds that might be deviating more from the Bavarian images, will reduce the initial (warm-start) performance, making it even more important to train the model actively with images from those new camera stations.

While the methods are indeed easy to use, it is helpful to have access to sufficient computing resources. 
Although it is possible to use the methods on a local machine with only CPUs, working on a machine with a GPU is substantially faster. A rather time-consuming step is the application of the \textit{Megadetector} for finding the bounding boxes, which, however, needs to be carried out only once before starting the active learning loop. There are attempts to approximate the \textit{Megadetector} with a far smaller object detection method \citep[e.g.,][]{rigoudy_deepfaune_2022} saving computing time. We still opt for the \textit{Megadetector} due to its very convincing performance. The other time-consuming step is the repeated tuning of the hyperparameters which can slow down the progress in the active learning loop; we, therefore, allow to optionally skip the tuning inside the active learning.
In many applications, though, human labor will remain the scarcest resource, and the need for it is greatly alleviated with our pipeline.

\paragraph{Future work}
There are some directions of future work that we would like to pursue:
\begin{enumerate*}
    \item Labeling bounding boxes instead of original images could further improve the performance of the models.
    \item More complex acquisition functions in the active learning pipeline could improve efficiency.
    \item We already have first results on incorporating methods from explainable AI in the pipeline. Explaining the predictions of the model can help to find systematic errors.
    \item Quantifying the predictive uncertainty of the model and the derived population count might give more realistic insights into the data. 
    \item Using metadata such as time, date, etc.\ as additional input has the potential to further improve the results. 
    \item Combining machine learning and domain knowledge could be a promising next step, see, e.g., \cite{tuia_perspectives_2022}.
    \item Finally, we plan on developing an R package that helps researchers with downstream analyses of the information from the image classification pipeline.
\end{enumerate*}

\newpage

\section*{Acknowledgements}

We thank Michael Jeske for helping with labeling the images. We thank Holger Löwe for helping with visualizations. 

\section*{Funding}

CB, WP, and AM received funding within the project LandKlif, funded by the Bavarian State Ministry of Science and the Arts in the context of the Bavarian Climate Research Network (bayklif). HE, WP, and CB received funding from the Bavarian State Ministry of Agriculture and Forestry (grant number ST375). TW received funding from the Deutsche Forschungsgemeinschaft (DFG, German Research Foundation) as part of BERD@NFDI (grant number 460037581).
LW received funding from the DAAD programme Konrad Zuse Schools of Excellence in Artificial Intelligence, sponsored by the Federal Ministry of Education and Research, Germany.

\section*{Conflict of Interest statement}

The authors declare no conflicts of interest.

\section*{Author Contributions}

Conceptualization, LB, HE, WP, CB, and AM;
Methodology, LB, OC, LW, and TW;
Software, LB, OC, LW, and TW;
Formal analysis, LB, OC, and LW;
Resources, WP and AM;
Data curation, CB, HE and HN;
Writing - Original Draft, LB;
Writing - Review \& Editing, all authors;
Visualization, OC and LW

\section*{Data Availability}

The image data set will be archived on Dryad upon acceptance. 

\bibliography{mybib-manually} 

\newpage
\appendix

\section{Additional results on different data}
\label{sec:app}

To investigate the behavior of our proposed method on data from a different study site, we analyzed images from the Channel Islands Camera Traps \citep{channel_island_2021} obtained via the LILA database.\footnote{https://lila.science/datasets/channel-islands-camera-traps/} We used a subset of 50,868 images (from 7 out of 73 camera stations) to make the size comparable to the Bavarian data analyzed above and also split the images into in-sample and out-of-sample data by camera station (3 and 4 stations for in-sample and out-of-sample, respectively). The dataset comprises 4 animal classes as well as an ``empty'' and an ``other'' class, where we removed the latter because it contains just a minimal amount of images (less than $0.14 \%$). Table \ref{tab:train_data_channel} summarizes the distribution of classes in the images used here.

\begin{table}[H]
\centering
\caption{Number of images per class (Channel Islands).}
\begin{tabular}{lrrr}
\label{tab:train_data_channel}
Species & in-sample & out-of-sample & total \\
\midrule
fox & 4,147 & 12,462 & 16,609\\
skunk     &37 &   142 & 179\\
rodent      &1,397 &  483 & 1,880\\
bird & 210 &  568 & 778\\
empty       &14,445 & 16,977 &  31,422\\
\midrule
total & 20,236 & 30,632 & 50,868\\
\bottomrule
\end{tabular}
\end{table}

\begin{figure}[H]
\centering
\begin{fig_publish}
\includegraphics[width=.7\linewidth]{fig/cm_insample_ci.eps}
\end{fig_publish}
\begin{fig_count_words}
\includegraphics[width=.7\linewidth]{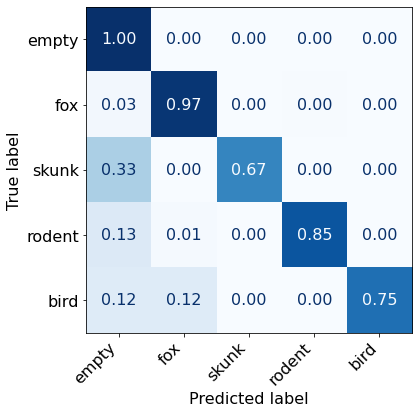}
\end{fig_count_words}
\caption{In-sample performance (Channel Islands)}
\label{fig:multi_conf_channel}
\end{figure}

We applied the same tuning strategy as explained above in Section \ref{sec:res_ins} and show the results in Table \ref{tab:tuningmdv5_channel}. The best hyperparameter combination is again Xception with a confidence threshold of $\alpha=0.1$ and as above, high thresholds of $\alpha=0.9$ perform worst. We can also see that the model is performing better than for the data from Bavaria, see Figure \ref{fig:multi_conf_channel} for detailed results on the test data (in-sample). As before, performance for larger classes (fox, rodent) is higher than for classes with fewer images (skunk, bird).

\begin{table}[H]
\centering
\caption{Hyperparameter tuning: best and worst configurations sorted by F1-score (MDv5, Channel Islands)}
\begin{tabular}{lrr}
\label{tab:tuningmdv5_channel}
Confidence & Architecture & F1-score\\
\midrule
\textbf{0.1} & \textbf{Xception} & \textbf{0.974} \\
0.3 & Xception & 0.973 \\
0.1 & InceptionResNetV2 & 0.965 \\
0.5 & Xception & 0.964 \\
$\vdots$ & $\vdots$ & $\vdots$ \\
0.9 & InceptionResNetV2 & 0.814 \\
0.9 & DenseNet121 & 0.814 \\
0.9 & Xception & 0.813 \\
\bottomrule
\end{tabular}
\end{table}

Table \ref{tab:empty_resultsmdv5_channel} shows the performance of the pipeline in the binary classification of empty vs. non-empty for different thresholds and using Xception, also underlining the stronger performance for smaller thresholds.

\begin{table}[H]
\centering
\caption{Empty vs. non-empty images (MDv5, Channel Islands)}
\begin{tabular}{l|c|ccc|ccc}
\label{tab:empty_resultsmdv5_channel}
& & & empty & & & non-empty & \\
\hline
Confidence & Accuracy & Precision & Recall & F1 & Precision & Recall & F1 \\[.1ex]
\hline
0.1 & 0.982	& 0.980 & 0.995 & 0.988 & 0.987 & 0.949 & 0.968 \\
0.3 & 0.981	& 0.978	& 0.995	& 0.987	& 0.988	& 0.943	& 0.965 \\
0.5 & 0.974	& 0.966	& 0.999	& 0.982	& 0.997	& 0.911	& 0.952  \\
0.7 & 0.937 & 0.919 & 1.000 & 0.958 & 1.000 & 0.778 & 0.875 \\
0.9 & 0.862 & 0.839 & 1.000 & 0.912 & 1.000 & 0.514 & 0.679 \\
\bottomrule
\end{tabular}
\end{table}

\begin{figure}[H]
\centering
\begin{fig_publish}
\includegraphics[width=.7\linewidth]{fig/cm_oosample_ci.eps}
\end{fig_publish}
\begin{fig_count_words}
\includegraphics[width=.7\linewidth]{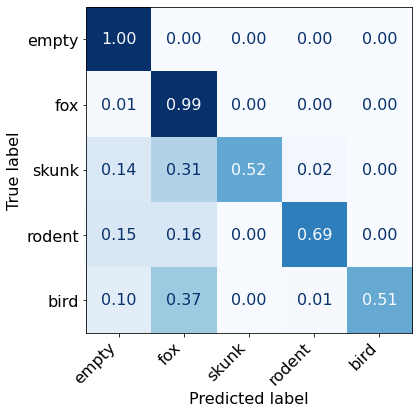}
\end{fig_count_words}
\caption{Out-of-sample performance without AL (Channel Islands).}
\label{fig:oos_multi_channel}
\end{figure}

\begin{figure}[H]
\centering
\begin{fig_publish}
\includegraphics[width=.7\linewidth]{fig/cm_active_ci.eps}
\end{fig_publish}
\begin{fig_count_words}
\includegraphics[width=.7\linewidth]{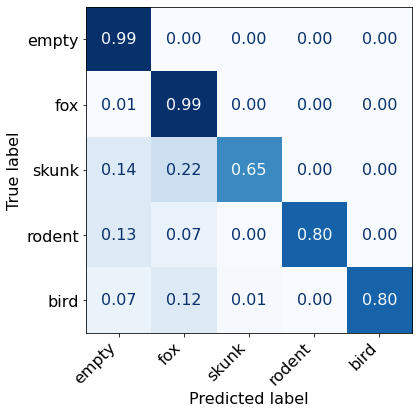}
\end{fig_count_words}
\caption{Out-of-sample performance with AL using $29\%$ of available data (Channel Islands).}
\label{fig:al_multi_channel}
\end{figure}

Table \ref{tab:ins-oos_channel} compares the performance on in-sample (first row) and out-of-sample (second row), showing only a minimal drop in performance when moving from in-sample to out-of-sample (see also Figure \ref{fig:oos_multi_channel} for detailed out-of-sample results on the test set). The active learning pipeline again improves the results (third row) and achieves 99.6\% of the performance compared to using the entire out-of-sample training data (fourth row) by just training on 3,968 images, i.e., 29\% of the images. The higher performance of AL-100$\%$ compared to in-sample suggests that the out-of-sample data poses an easier challenge for the specific case of the Channel Islands data. The largest class (fox) has fewer ``false-empty'' predictions in out-of-sample (Figure \ref{fig:oos_multi_channel}) compared to in-sample (Figure \ref{fig:multi_conf_channel}) and the overall drop in performance (Table \ref{tab:ins-oos_channel}, first and second row) stems from the smaller classes. Using AL, the performance on these smaller classes increases (Figure \ref{fig:al_multi_channel}).

\begin{table}[H]
\centering
\caption{Multi-class classification (MDv5, Channel Islands). Precision and F1-score are weighted averages of 5 classes.}
\begin{tabular}{lrrrr}
\label{tab:ins-oos_channel}
{} & Accuracy & Precision &  F1\\
\midrule
In-sample &  0.979 & 0.979 &  0.979 \\
Out-of-sample & 0.978 & 0.977  & 0.976 \\
Active learning -- 29\% & 0.983 & 0.983 &  0.983 \\
Active learning -- 100\%& 0.987 & 0.986 &  0.986 \\
\bottomrule
\end{tabular}
\end{table}

\begin{figure}[H]
\centering
\begin{fig_publish}
\includegraphics[width=\linewidth]{fig/active_learning_ci.eps}
\end{fig_publish}
\begin{fig_count_words}
\includegraphics[width=\linewidth]{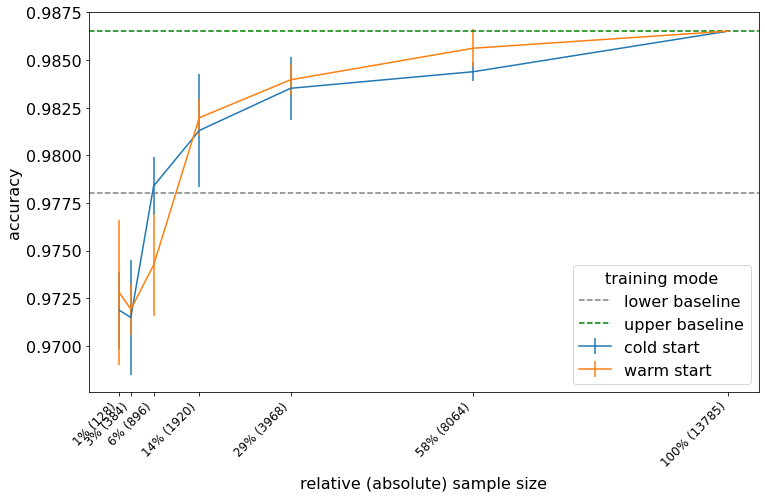}
\end{fig_count_words}
\caption{Active-learning performance (5-class accuracy, Channel Islands) for cold and warm start as compared to out-of-sample prediction (lower horizontal line) and training on the entire data set (upper horizontal line). Error bars indicate variability over 3 different random seeds.}
\label{fig:oos_active_channel}
\end{figure}

Figure \ref{fig:oos_active_channel} shows how quickly the performance increases while iteratively including more labeled images for training in the active learning loop. 
We can see that accuracy surpasses the lower baseline and reaches a close-to-optimal level early on.
Again, the algorithm achieves better performance overall on the Channel Islands data compared to the Bavarian data. 
In this case, warm-starting does not improve the performance substantially, compared to cold-starting.
Finally, Figure \ref{fig:al_multi_channel} shows detailed classification results for active learning after using 3,968 images for training.
While performance on the larger classes is high for out-of-sample as well, the active learning procedure benefits the smaller classes, identifying 25.0\% more skunks, 15.9\% more rodents, and 56.9\% more birds correctly. 

\end{document}